%% file: main.tex
\PassOptionsToPackage{numbers,compress}{natbib}
\documentclass{article}
\usepackage[utf8]{inputenc}
\usepackage[preprint]{neurips_2026}
\usepackage{microtype}
\usepackage{graphicx}
\usepackage{amsmath}
\usepackage{amssymb}
\usepackage{booktabs}
\usepackage{multirow}
\usepackage{longtable}
\usepackage{float}
\usepackage{placeins}
\usepackage{needspace}
\usepackage{hyperref}
\usepackage{xurl}
\usepackage{xcolor}

\hypersetup{
  hidelinks,
  colorlinks=false,
  pdfauthor={Gabriel Garcia},
  pdftitle={Protection Is (Nearly) All You Need: Structural Protection Dominates Scoring in Globally Capped KV Eviction},
}

\newcommand{\FaithAdaF}{93}      
\newcommand{\FaithQuestF}{92}    
\newcommand{\FaithAdaFone}{0.167}  
\newcommand{\FaithQuestFone}{0.164} 
\newcommand{\NIAHFullcacheF}{0.117}  
\newcommand{\NIAHLRUPct}{30}     
\newcommand{\NIAHHOPct}{22}      
\newcommand{\NIAHRndPct}{27}     

\newcommand{\RecoveryCoreSeven}{69--90\%}   
\newcommand{\RecoveryTenPanel}{68--98\%}   

\newcommand{\PublicRepoURL}{https://github.com/gpgabriel25/KVCacheBoundaryProtection}

\title{Protection Is (Nearly) All You Need:\\Structural Protection Dominates Scoring in Globally Capped KV Eviction}
\author{Gabriel Garcia \\
Independent Researcher \\
\texttt{gpgabriel25@gmail.com}}
\date{}

\begin{document}
\raggedbottom
\widowpenalty=10000
\clubpenalty=10000

\maketitle

\begin{abstract}
We study KV cache eviction under a shared globally capped decode-time harness.
Seven policies (LRU, H2O, SnapKV, StreamingLLM, Ada-KV, QUEST, Random) share a prompt-boundary vulnerability: without structural protection, they collapse to near-zero quality on six pure-transformer models (F1$\leq$0.064).
Reserving 10\% of cache at each boundary recovers \RecoveryCoreSeven{} of the $C{=}2{,}048$ reference-ceiling quality on seven LongBench models at $C{=}256$ (13\% retention); a ten-model panel spans \RecoveryTenPanel{}.
An attention-mass pilot (Qwen2.5-3B, $N{=}30$) suggests why: the position-0 sink holds ${\sim}75\%$ of prefix mass, while other boundary tokens sit near ${\sim}0.41{\times}$ uniform expectation, so attention scorers retain the sink but still drop structurally critical tokens.
With protection, simplified score-isolation variants are TOST-equivalent to LRU at $K{=}32$ ($\Delta{=}0.02$); at $K{=}8$, attention policies pairwise converge yet beat LRU by 0.011--0.021 F1 across $C{=}256$ and $C{=}512$.
Faithful Ada-KV/QUEST add ${\sim}0.03$--$0.04$ F1 on Mistral-7B and Phi-3.5 beyond simplified variants.
A NIAH-32K regime-transfer pilot on Qwen3-4B (decode vs.\ prefill, $C{\in}\{512,2048\}$) shows near-identical protection lifts (ratio 0.99--1.00).
At 64K, protection helps but recovery is modest; faithful per-head scoring matches full-cache ceiling on Gemma-3-4B at 6.3\% retention only when the model already supports strong 64K retrieval without eviction.
Overall: protection dominates; scoring differences are secondary once boundaries are guarded; per-head allocation gives a further modest gain.
\end{abstract}

\section{Introduction}
\label{sec:intro}

Transformer-based language models cache key-value (KV) representations during autoregressive generation to avoid recomputation~\citep{vaswani2017attention}.
As context lengths scale, KV cache memory grows linearly, making explicit cache management necessary under memory constraints~\citep{pope2023efficiently}.

A substantial body of work has developed increasingly sophisticated eviction heuristics.
H2O~\citep{zhang2023h2o} retains positions receiving the highest cumulative attention mass.
SnapKV~\citep{ge2024snapkv} combines attention mass with current-query relevance via observation windows.
StreamingLLM~\citep{xiao2023streamingllm} protects attention-sink tokens and maintains a sliding window.
Scissorhands~\citep{liu2023scissorhands} exploits persistence of importance across decoding steps.
Ada-KV~\citep{feng2024adakv} introduces adaptive per-head budget allocation.
QUEST~\citep{tang2024quest} adds query-aware sparsity to eviction scoring.
Each proposes a different scoring function and evaluates against different baselines, creating an impression of meaningful performance differences among eviction algorithms.

In a shared decode-time, globally capped evaluation regime, much of the reported spread among eviction policies is a shared confound; scoring still matters on architectures with sufficient head diversity, but as a second-order effect.
Several of these methods embed implicit forms of structural protection: StreamingLLM explicitly retains sink tokens, and attention-based scoring in H2O and SnapKV incidentally favours boundary positions, but no prior work isolates or quantifies the protection effect.
When we evaluate all seven policies without structural protection of prompt-boundary positions in a common global-cap harness, all produce near-zero output quality (F1$\leq$0.064) under aggressive cache compression.
The catastrophic failure is not caused by poor scoring; it is caused by evicting structurally critical positions (instruction delimiters, question boundaries, attention sinks) that anchor coherent generation.
This globally capped decode-time regime, where a single capacity budget is shared across all heads and eviction occurs token-by-token during generation, appears in the evaluation protocols of H2O~\citep{zhang2023h2o}, SnapKV~\citep{ge2024snapkv}, Scissorhands~\citep{liu2023scissorhands}, and parts of the Ada-KV~\citep{feng2024adakv} and QUEST~\citep{tang2024quest} evaluation suites (with minor variations across papers).
This paper is a deconfounding study of that shared harness.
To confirm that the harness we study matches the cited papers, we conducted a code-level audit of the canonical reference implementations (H2O, SnapKV, Ada-KV, QUEST, vLLM; detailed in \S\ref{para:ref_impl_audit}), which finds that (i)~all four eviction implementations retain a recent-token window but (ii)~none implements structural first-token sink protection by position; the vulnerability we identify is therefore a real property of published reference code rather than a harness artifact.

Our contribution is a systematic empirical deconfounding study: we isolate the structural protection effect from the scoring-heuristic effect and show that the former dominates in the global-cap regime we study:

\begin{enumerate}
    \item \textbf{Structural vulnerability identification.} We demonstrate that seven representative eviction policies (LRU, H2O, SnapKV, StreamingLLM, Ada-KV, QUEST, and Random) all collapse catastrophically without structural protection, producing degenerate output across all tested cache capacities (Section~\ref{sec:vulnerability}).

    \item \textbf{Robust fix via structural protection.} Protecting 10\% of cache capacity at each end (prefix + suffix) transforms all tested policies from catastrophic failure to near-ceiling performance: 79--92\% of full-cache quality at $C{=}256$ on Qwen2.5-3B (Table~\ref{tab:universality}), with \RecoveryCoreSeven{} across the seven-model LongBench panel at the same budget (Section~\ref{sec:crossmodel}, Table~\ref{tab:3model}).

    \item \textbf{Head-count-dependent policy effects under protection.} Under protection in the decode-time global-cap harness, the strongest empirical convergence evidence comes from $K{=}8$ and $K{=}32$: at $K{=}8$, the four attention-based policies are pairwise equivalent but collectively outperform LRU by 0.011--0.021 F1 across $C{=}256$ and $C{=}512$, while at $K{=}32$ the simplified/common-harness variants of LRU, H2O, SnapKV, Ada-KV, and QUEST are TOST-equivalent at $\Delta{=}0.02$ ($p{<}0.05$). The $K{=}2$ case is a low-head-diversity sanity check rather than the core evidence: in our realized common-harness traces, the attention-based scores induce the same eviction behavior, but we do not use this as the main proof of policy convergence. Faithful per-head reproductions of Ada-KV and QUEST recover near-ceiling quality on both Phi-3.5 ($K{=}32$; \FaithAdaF{}\% and \FaithQuestF{}\% of full-cache F1) and Mistral-7B ($K{=}8$; 95\% and 94\%), versus ${\sim}79$--$80\%$ for simplified global-ranking variants; faithful H2O and SnapKV similarly converge on Qwen2.5-3B ($K{=}2$) and Mistral-7B ($K{=}8$). Per-head selection provides a genuine benefit (${\sim}0.03$--$0.04$ F1 absolute) but this gap is modest relative to the protection lift (${\sim}0.11$--$0.14$ F1), confirming that structural protection remains the primary quality factor (Sections~\ref{sec:equivalence}, Appendix~\ref{app:faithfulness}).

    \item \textbf{Protection sensitivity curve.} A sweep over five protection levels (0\%--20\%) identifies 10\% as the empirical sweet spot: 5\% captures 78\% of ceiling, 10\% captures 89\%, and further increases provide no significant improvement (Section~\ref{sec:sensitivity}).

    \item \textbf{Negative result on learned credit.} In the original paper-direction experiment (online-credit v2), a 4-feature linear counterfactual estimator trained during inference adds no measurable benefit over LRU+protection across five tested capacities ($C \in \{64, 96, 128, 256, 512\}$), including extreme compression (Section~\ref{sec:credit_negative}).

    \item \textbf{Cross-architecture evidence.} We replicate the main protection-first pattern across seven core models spanning 1.5B--27B parameters, three attention architectures (GQA, full MHA, DeltaNet-GQA hybrid), and four model families, then add a three-model current-generation sweep for ten evaluated models total (Section~\ref{sec:crossmodel}).
    On a synthetic Needle-in-a-Haystack benchmark, the protection effect replicates: unprotected eviction destroys retrieval (F1$\leq$0.007) while protection provides consistent relative improvement with minimal cross-policy spread (Appendix~\ref{sec:niah}).
    The three-part pattern further holds on 48 LongBench items at 11K tokens, 32K-token NarrativeQA contexts, 40 Multi-News summarization items, and under temperature sampling ($T{=}0.7$).

    \item \textbf{Pure-transformer 64K stress test.} At 64K-token needle-in-a-haystack retrieval on pure-transformer Qwen3-4B-Instruct-2507, protected LRU remains the strongest low-retention condition: LRU+prot at $C{=}4{,}096$ recovers 39.8\% of ceiling versus 34.7\% without protection. At matched budget ($C{=}4{,}096$), H2O+prot reaches only 7.8\% of ceiling and H2O without protection collapses to 0.000 F1; even at double the budget ($C{=}8{,}192$), H2O+prot and Random+prot recover only 12.4\% and 14.9\%. Recovery is weaker than at shorter contexts but protection remains a positive lever (Section~\ref{sec:64k}).\end{enumerate}

These findings have immediate practical value: adding prefix/suffix guards to an eviction implementation (a set-membership check during candidate selection, negligible algorithmic overhead) recovers near-optimal quality in the tested harness.
At the systems level, we observe that nnx-level JIT compilation in this regime improves raw throughput by two orders of magnitude (from 0.09 to 43.6 tok/s on Qwen3-14B, v6e-4 TPU), confirming that the protection mechanism carries no throughput penalty beyond standard eager-to-compiled acceleration.
The claims are threefold: catastrophic failure without structural guards, broad convergence among simplified scorers once guards are added, and a smaller but real lift from faithful per-head allocation.

\section{Related Work}
\label{sec:related}

\paragraph{KV cache eviction.}
StreamingLLM~\citep{xiao2023streamingllm} retains attention-sink tokens plus a sliding window for stable streaming inference.
H2O~\citep{zhang2023h2o} retains ``heavy-hitter'' tokens receiving disproportionate cumulative attention mass.
Scissorhands~\citep{liu2023scissorhands} exploits the persistence-of-importance hypothesis.
SnapKV~\citep{ge2024snapkv} identifies important positions before generation via observation windows.
QUEST~\citep{tang2024quest} adds query-aware sparsity.
TrimKV~\citep{li2025trimkv} uses layer-aware token merging.
ForesightKV~\citep{zhang2025foresightkv} estimates attention scores by prediction.
Ada-KV~\citep{feng2024adakv} introduces head-wise adaptive budget allocation, optimizing how eviction budgets are distributed across attention heads rather than which tokens to evict.
PyramidInfer~\citep{yang2024pyramidinfer} compresses the KV cache layer-wise by exploiting the observation that the number of influential positions decreases in deeper layers.
InfLLM~\citep{xiao2024infllm} offloads evicted KV entries to CPU memory for on-demand retrieval, coupling eviction with an external memory buffer.
Keyformer~\citep{adnan2024keyformer} identifies key tokens whose retention suffices for near-lossless generation.
RazorAttention~\citep{tang2024razorattention} compresses non-retrieval heads aggressively while preserving retrieval heads, effectively applying structural protection at the head granularity.
Many of these methods include mechanisms that can act as implicit structural protection (attention-sink retention, observation windows, retrieval-head preservation, or similar), but none isolates that contribution under a common decode-time harness with and without explicit boundary guards.
Our work deconfounds the scoring-heuristic effect from the structural-protection effect by evaluating all policies under a common global-cap harness with and without explicit boundary guards.

\paragraph{KV cache quantization.}
KIVI~\citep{liu2024kivi} and KVQuant~\citep{hooper2024kvquant} apply per-channel or per-token quantization to compress cache storage.
MiKV~\citep{liu2024mikv} combines head-level importance with mixed-precision storage.
These methods are orthogonal to eviction policy and do not address the structural vulnerability we identify.

\paragraph{Efficient inference systems.}
PagedAttention~\citep{kwon2023vllm} virtualizes KV memory for efficient multi-tenant serving.
FlashAttention~\citep{dao2022flashattention,dao2023flashattention2} provides IO-aware attention kernels.
Our structural protection is compatible with these systems-level optimizations.

\paragraph{Learned caching.}
The algorithms-with-predictions framework~\citep{lykouris2021competitive} studies how machine-learned advice can improve classical caching algorithms.
Our credit learning system implements this paradigm online, providing an empirical test of whether learned eviction decisions can improve over simple heuristics once the structural confound is removed.
Our finding that they cannot is informative for this research direction.

\section{Experimental Setup}
\label{sec:setup}

\paragraph{Models.}
We evaluate ten models spanning four vendor families in the seven-model LongBench panel (Alibaba/Qwen, Microsoft/Phi, Mistral, 01.AI/Yi) and five families in the full ten-model panel (adding Google/Gemma); see Table~\ref{tab:models} and~\citep{qwen2024qwen25,qwen2026qwen35,qwen2025qwen3,abdin2024phi3,abdin2025phi4mini,jiang2023mistral,young2024yi,gemma2025gemma3}.\footnote{Qwen3.5-27B uses the text-only inference path of the multimodal checkpoint. Qwen3-4B-Instruct-2507 is the pure-transformer 64K control.}
\begin{itemize}
    \item \textbf{Qwen2.5-3B-Instruct} (36 layers, 2048-dim, GQA with 16 query heads and 2 KV heads)
    \item \textbf{Qwen2.5-7B-Instruct} (scale validation; 28 layers, 3584-dim, GQA with 28 query heads and 4 KV heads)
    \item \textbf{Qwen3.5-27B} (HuggingFace id \texttt{Qwen/Qwen3.5-27B}; text-only inference path; hybrid architecture validation; 64 layers, 5120-dim, 48 DeltaNet linear-attention layers + 16 Gated Attention layers with 24 query heads and 4 KV heads; 27B parameters)
    \item \textbf{Qwen2.5-1.5B-Instruct} (same-family validation; 28 layers, 1536-dim, GQA with 12 query heads and 2 KV heads)
    \item \textbf{Phi-3.5-mini-instruct} (cross-architecture validation; 32 layers, 3072-dim, full multi-head attention with 32 query heads and 32 KV heads; 3.8B parameters)
    \item \textbf{Mistral-7B-Instruct-v0.3} (cross-family validation; 32 layers, 4096-dim, GQA with 32 query heads and 8 KV heads; 7.2B parameters)
    \item \textbf{Yi-1.5-6B-Chat} (fourth-family cross-validation; 32 layers, 4096-dim, GQA with 32 query heads and 4 KV heads; 6.0B parameters~\citep{young2024yi})
    \item \textbf{Qwen3-4B-Instruct-2507}~\citep{qwen2025qwen3} (current-generation Alibaba dense transformer, 4B class)
    \item \textbf{Phi-4-mini-instruct}~\citep{abdin2025phi4mini} (current-generation Microsoft 4B-class instruct model)
    \item \textbf{Gemma-3-4B-IT}~\citep{gemma2025gemma3} (current-generation Google 4B-class instruct model)
\end{itemize}
The Qwen2.5 models use grouped-query attention (GQA) with 2--4 KV heads; Phi-3.5 uses standard multi-head attention (MHA) with 32 KV heads; Mistral uses GQA with 8 KV heads; Yi-1.5 uses GQA with 4 KV heads; and Qwen3.5-27B uses a hybrid architecture where 75\% of layers are DeltaNet (linear attention with recurrent state) and 25\% are Gated Attention layers (24 query heads, 4 KV heads), providing a test of whether the protection effect extends beyond pure-transformer models.

\paragraph{Evidentiary roles.}
Core protection-effect evidence comes from the Qwen2.5 LongBench matrix, where all seven policies and multiple capacities are swept on the same balanced $N{=}162$ item set.
Architecture-diversity evidence comes from Phi-3.5, Mistral-7B, Yi-1.5, and Qwen3.5-27B, which test full-MHA, higher-$K$ GQA, fourth-family GQA, and hybrid DeltaNet+GQA settings.
Long-context stress evidence comes from Qwen3-4B-Instruct-2507, Phi-4-mini, and Gemma-3-4B-IT on synthetic 64K retrieval, where the relevant question is not only which eviction policy is best but whether the model has a non-trivial full-cache retrieval ceiling.
This combination tests whether our findings generalize across attention design, model scale, model family, and architectural paradigm; in particular, Phi-3.5 ($K{=}32$) and Mistral-7B ($K{=}8$), where scoring heuristics produce genuinely distinct orderings across heads, provide the strongest tests of policy convergence (Section~\ref{sec:crossmodel}), while Qwen3.5-27B tests generalization to a fundamentally different layer composition.
Generation uses greedy decoding with a maximum of 128 new tokens.

\begin{table}[H]
\centering
\small
\caption{Evaluated model checkpoints (Hugging Face ids, native context limits, licenses). All LongBench QA runs use the instruct/chat variants below unless noted.}
\label{tab:models}
\begin{tabular}{@{}llrl@{}}
\toprule
Paper name & Hugging Face checkpoint & Context & License \\
\midrule
Qwen2.5-3B-Instruct & \texttt{Qwen/Qwen2.5-3B-Instruct} & 32K & Apache-2.0 \\
Qwen2.5-7B-Instruct & \texttt{Qwen/Qwen2.5-7B-Instruct} & 128K & Apache-2.0 \\
Qwen2.5-1.5B-Instruct & \texttt{Qwen/Qwen2.5-1.5B-Instruct} & 32K & Apache-2.0 \\
Phi-3.5-mini-instruct & \texttt{microsoft/Phi-3.5-mini-instruct} & 128K & MIT \\
Mistral-7B-Instruct-v0.3 & \texttt{mistralai/Mistral-7B-Instruct-v0.3} & 32K & Apache-2.0 \\
Yi-1.5-6B-Chat & \texttt{01-ai/Yi-1.5-6B-Chat} & 4K & Apache-2.0 \\
Qwen3.5-27B & \texttt{Qwen/Qwen3.5-27B} & 262K & Apache-2.0 \\
Qwen3-4B-Instruct-2507 (Qwen3-4B) & \texttt{Qwen/Qwen3-4B-Instruct-2507} & 262K & Apache-2.0 \\
Phi-4-mini & \texttt{microsoft/Phi-4-mini-instruct} & 128K & MIT \\
Gemma-3-4B-IT & \texttt{google/gemma-3-4b-it} & 128K & Gemma \\
\bottomrule
\end{tabular}
\end{table}

\paragraph{Benchmark.}
We use a balanced subset of LongBench~\citep{bai2023longbench} comprising $N{=}162$ question-answer items from six subtasks: 2WikiMQA~(40), Qasper~(40), MultifieldQA-EN~(40), HotPotQA~(30), NarrativeQA~(8), and MuSiQue~(4).
Contexts range from 549 to 5{,}996 words in the raw corpus ($\sim$1{,}920 tokens on average after tokenization).
Quality is measured by token-level F1, scored as the maximum over all reference answers per item (standard LongBench protocol).

\paragraph{Reference ceiling ($C{=}2{,}048$).}
All primary LongBench QA runs tokenize prompts with a hard cap of 1{,}920 tokens (\texttt{max\_cache\_len}{=}2{,}048 minus 128 generation slots); longer inputs are truncated to fit before prefill, so eviction never sees prompts above this budget.
Headline comparisons use $C{=}2{,}048$ with capacity $\geq$ the prefilled length and no decode-time eviction as the \emph{reference ceiling} under this truncation policy (not an unlimited-context oracle).
Recovery percentages are relative to this reference ceiling measured under identical tokenization.
We report 95\% bootstrap confidence intervals (10{,}000 resamples) and paired Wilcoxon signed-rank tests for pairwise comparisons.
For families of convergence tests (Table~\ref{tab:equivalence}), we apply Holm--Bonferroni correction~\citep{holm1979simple}; all reported non-significant results remain non-significant after correction.
For the policy-convergence claim, we supplement non-significance with the Two One-Sided Tests (TOST) procedure~\citep{schuirmann1987tost,lakens2017equivalence} using a pre-specified practical equivalence margin of $\Delta{=}0.02$ absolute F1, a fixed threshold independent of model ceiling, chosen as ${\sim}6$--$11\%$ of ceiling depending on the model, well above measurement noise yet small enough to be practically meaningful (Section~\ref{sec:equivalence}).

\paragraph{KV-coupled evaluation (``globally capped decode-time'' regime).}
All eviction policies physically modify the KV cache during autoregressive decode:
(1)~the full context is prefilled, creating the complete KV cache;
(2)~the policy reduces the cache from context length to target capacity~$C$;
(3)~every $\tau{=}8$ decode steps, the policy is re-invoked to maintain capacity as new tokens are generated.
The model generates with the physically modified cache; evicted positions contribute nothing to attention.
We term this the \emph{globally capped decode-time} regime: a single global capacity budget $C$ is enforced across all heads and layers, eviction occurs during decode (not prefill), and evicted entries are permanently removed.
This regime matches recent KV-eviction evaluations~\citep{zhang2023h2o,ge2024snapkv,tang2024quest,feng2024adakv}; we adapt methods originally designed for prefill-time or per-head budgets into a common decode-time harness for deconfounding (Appendix~\ref{app:faithfulness}).
It differs from production serving paradigms such as per-request PagedAttention~\citep{kwon2023vllm} or streaming prefill (Section~\ref{sec:discussion}).
We add one matched prefill-time stress test at 32K context as limited transfer evidence, but it is not a substitute for a native serving-system evaluation.

\paragraph{Infrastructure.}
Experiments run on Cloud TPU v5e-64 (16 workers, single-chip mode) and v6e-4 (four-chip tensor-parallel decode) with JAX 0.6.2--0.9.x.
Model forward passes in our TPU experiments use JIT-compiled decode.
Appendix~\ref{app:p99} reports per-decode-step latency from the real JAX forward path in \texttt{KVCoupledQwen35Generator} (Table~\ref{tab:p99_jax}).

\paragraph{Policies.}
We evaluate seven eviction policies spanning the full spectrum from purely random to sophisticated attention-aware scoring:
\begin{itemize}
    \item \textbf{LRU}: evicts the least recently accessed position.
    \item \textbf{H2O}~\citep{zhang2023h2o}: retains positions by cumulative attention mass.
    \item \textbf{SnapKV}~\citep{ge2024snapkv}: combines attention mass with current attention score.
    \item \textbf{StreamingLLM-style sink+window (SLW)}~\citep{xiao2023streamingllm}: retains the method's native sink tokens ($n_{\text{sink}}{=}4$) plus a recency window; we abbreviate this policy as \texttt{SLW} in tables and figures. The \texttt{+prot} suffix adds our bilateral prefix/suffix guard on top of the native sink/window rule, which is distinct from StreamingLLM's original streaming-prefill deployment.
    \item \textbf{Ada-KV}~\citep{feng2024adakv}: scores by cumulative attention plus half the maximum single-step attention received, with head-wise adaptive budget allocation.
    \item \textbf{QUEST}~\citep{tang2024quest}: scores by current-step attention only, providing a query-aware importance estimate.
    \item \textbf{Random}: selects eviction victims uniformly at random from the unprotected region (no scoring information).
\end{itemize}
The $C{=}2{,}048$ reference ceiling (no decode-time eviction; prompts truncated as above) achieves F1$=$0.315 (Qwen2.5-3B), F1$=$0.313 (Qwen2.5-7B), F1$=$0.304 (Qwen2.5-1.5B), F1$=$0.179 (Phi-3.5), and F1$=$0.234 (Mistral-7B).\footnote{These reference ceilings are lower than published LongBench leaderboard scores for comparable models because our evaluation uses a 128-token generation cap (vs.\ 512+ in official benchmarks), greedy decoding (vs.\ sampling or beam search), model-specific chat templates applied uniformly, and the 1{,}920-token prompt cap above.
The Phi-3.5-mini ceiling is notably low (F1$=$0.179), indicating weaker baseline performance on our balanced QA subset than on its official benchmark reports.
Since all comparisons are relative to the \emph{same} reference ceiling measured under identical conditions, the recovery percentages remain valid internal comparisons.}

\paragraph{Structural protection.}
When enabled, protection reserves $\lceil \rho \cdot C \rceil$ positions at each end of the cache (prefix and suffix), with a minimum of 4 positions per side, where $\rho$ is the protection fraction and $C$ is the cache capacity.
Protected positions are excluded from the evictable set; only unprotected positions are candidates for eviction.
The default protection fraction is $\rho{=}0.10$ (10\% prefix + 10\% suffix = 20\% of capacity reserved).

\paragraph{Online credit learning.}
To test whether learned eviction decisions improve over simple heuristics, we implement an online counterfactual credit estimator that learns during inference:
a 4-feature linear model observes log-probability deltas after eviction events and attributes credit to evicted positions via attention-weighted proportional assignment.
The system includes uncertainty gating (falls back to LRU when uncertain), progressive blending (linear ramp from 0 to 40\% blend), and noise thresholding.
This represents a principled attempt at learned caching in the algorithms-with-predictions framework.

\section{The Structural Vulnerability of KV Cache Eviction}
\label{sec:vulnerability}

Our central finding begins with a negative observation: \emph{all} tested eviction policies produce near-zero output quality under aggressive cache compression.

Table~\ref{tab:vulnerability} reports F1 scores for four policies at three representative capacities without structural protection.

\begin{table}[H]
    \centering
    \small
    \resizebox{\linewidth}{!}{%
    \begin{tabular}{lccc}
        \toprule
        Policy & $C{=}128$ (7\%) & $C{=}256$ (13\%) & $C{=}512$ (27\%) \\
        \midrule
        LRU         & $0.019 \pm 0.013$ & $0.011 \pm 0.012$ & $0.010 \pm 0.012$ \\
        H2O         & $0.030 \pm 0.017$ & $0.038 \pm 0.016$ & $0.038 \pm 0.011$ \\
        SnapKV      & $0.030 \pm 0.017$ & $0.038 \pm 0.015$ & $0.037 \pm 0.011$ \\
        SLW & $0.021 \pm 0.013$ & $0.027 \pm 0.013$ & $0.030 \pm 0.014$ \\
        \midrule
        \emph{Ref.\ ceiling} & \multicolumn{3}{c}{\emph{0.315 ($C{=}2{,}048$, no eviction)}} \\
        \bottomrule
    \end{tabular}
    }
    \caption{All eviction policies fail catastrophically without structural protection. F1 $\pm$ 95\% CI (Qwen2.5-3B, LongBench, $N{=}162$). \texttt{SLW} denotes StreamingLLM-style sink+window (native sink tokens plus recency window; Section~\ref{sec:setup}). Percentages indicate cache retention for $\sim$1{,}920-token prompts. The best unprotected policy (H2O at $C{=}256$) achieves only 12\% of the full-cache ceiling.}
    \label{tab:vulnerability}
\end{table}

The maximum F1 achieved by any unprotected policy at any capacity is 0.038 (H2O/SnapKV at $C{=}256\text{--}512$), just 12\% of the 0.315 full-cache ceiling.
LRU is worst (3.6\% of ceiling at $C{=}256$), and notably, performance does \emph{not} improve with larger cache sizes: LRU at $C{=}512$ (0.010) is actually lower than at $C{=}128$ (0.019), although the difference is not significant given overlapping confidence intervals.
This counterintuitive pattern is consistent with the phase-transition nature of the failure: once structural positions are evicted, output quality drops to near-zero regardless of how many middle-context tokens remain.
A larger unprotected cache simply accumulates more recently accessed middle-context tokens without preserving the critical prefix/suffix anchors, while at smaller capacities the recency window occasionally overlaps with question-boundary positions, yielding slightly higher (but still catastrophic) F1.

\begin{figure}[H]
    \centering
    \includegraphics[width=0.85\textwidth]{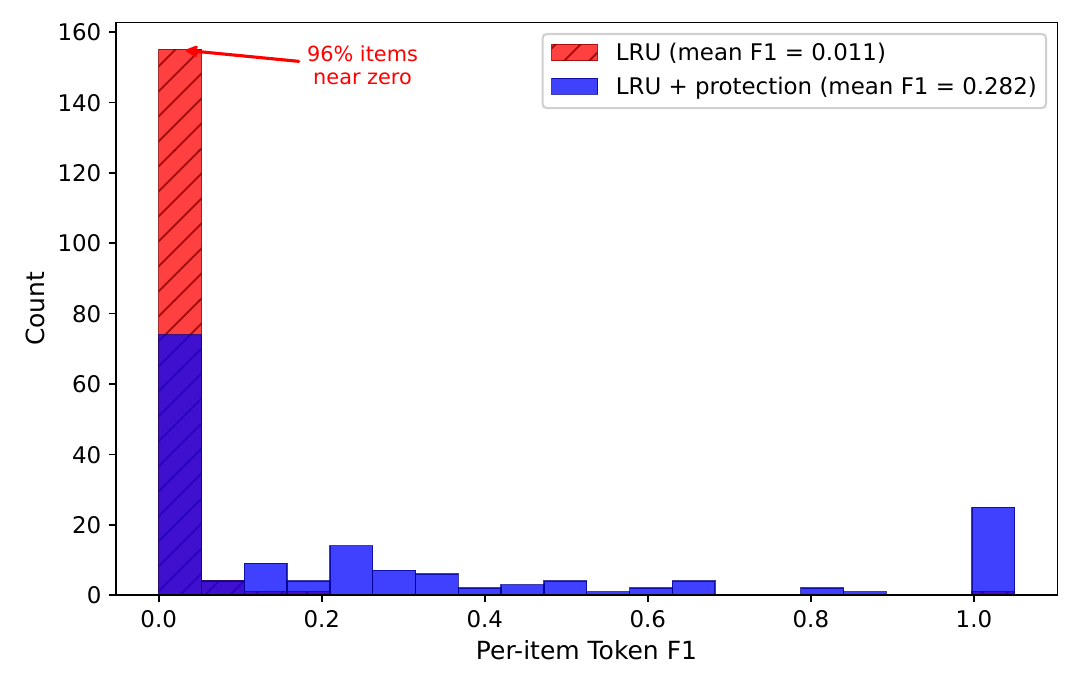}
    \caption{Per-item F1 distributions for LRU at $C{=}256$ (Qwen2.5-3B, $N{=}162$). Without protection (red), 96\% of items score near zero. Protection (blue) shifts roughly half the distribution to substantive F1, with 15\% of items recovering to $\geq$\,0.95. Of the 78 protected items still below F1\,$<$\,0.10, 83\% also score zero under the full cache, indicating model inability rather than a protection failure.}
    \label{fig:f1hist}
\end{figure}

\paragraph{Failure mechanism.}
Inspection of generation outputs reveals the cause: without protection, eviction removes prompt-boundary tokens (instruction delimiters, question markers, and attention sinks) that the model relies on for coherent generation.
Once these anchoring positions are evicted, the model enters repetition loops, producing token-maximal outputs (mean generation length 127.1 tokens out of 128 maximum) consisting of repeated fragments unrelated to the question.
A 4-gram repetition rate analysis confirms this is truly degenerate output, not merely wrong answers with low token overlap: unprotected LRU on Qwen2.5-3B at $C{=}256$ exhibits a mean 4-gram repetition rate of 95.3\%, compared to 0.1\% with protection; the pattern holds across models (Phi-3.5: 50.7\% vs.\ 0.8\%; Qwen2.5-7B: 14.8\% vs.\ 0.0\%) and policies (Random: 54.7\% vs.\ 0.4\%).
This is not a gradual quality degradation but a \emph{phase transition}: the model either generates coherently (with anchoring positions intact) or degenerately (without them).
Figure~\ref{fig:f1hist} visualises this dichotomy at the per-item level: without protection, 96\% of items score near zero, while adding protection shifts roughly half the distribution above F1\,$>$\,0.10 and recovers 15\% of items to near-perfect F1\,$\geq$\,0.95.

\paragraph{What the protected positions contain.}
While we do not instrument per-token eviction decisions, the identities of the protected positions follow deterministically from the prompt template and the protection rule.
Each model is evaluated with its \emph{native} instruct chat template via the tokenizer's built-in \texttt{apply\_chat\_template} path; across Qwen, Phi, and Mistral these templates all instantiate the same structural pattern of a system instruction, a user turn containing \texttt{Passage: ... Question: ...}, and an assistant-generation boundary.
At $C{=}256$ with 10\% protection (26 slots per side), the \emph{prefix guard} covers: the BOS token (known attention sink~\citep{xiao2023streamingllm}), the system-turn delimiter and instruction, and the first $\sim$10 passage tokens.
The \emph{suffix guard} covers: the final question tokens, the user-turn close delimiter, and the assistant-turn opening delimiter.
These are precisely the structural anchors that instruct-tuned models condition on: the system instruction defines the task, the question boundary specifies the target, and the turn delimiters segment the discourse.
Evicting any of these forces the model to attend only to mid-passage content tokens lacking any task or turn structure, explaining the degenerate repetition behaviour.
This architectural argument is consistent with the bilateral ablation (Table~\ref{tab:prefix_suffix}): prefix-only protection preserves the system instruction and attention sinks, suffix-only preserves the question boundary, and bilateral protection preserves both; each is independently necessary for quality recovery.

\paragraph{Universality of failure.}
The vulnerability is not specific to simple policies like LRU.
H2O's cumulative attention scoring, SnapKV's observation-window approach, the StreamingLLM-style sink+window rule (without our bilateral guard), Ada-KV's adaptive budget allocation~\citep{feng2024adakv}, QUEST's query-aware scoring~\citep{tang2024quest}, and even uniformly Random eviction all fail to preserve the critical positions.
While H2O and SnapKV slightly outperform LRU (0.038 vs.\ 0.011 at $C{=}256$), likely because cumulative decode-time scoring gradually accumulates higher eviction-resistance for high-frequency keys, the absolute quality remains catastrophic: at $C{=}256$, all seven policies achieve less than 9\% of ceiling; even at $C{=}512$, the best unprotected policy (H2O, 0.038) reaches only 12\% of ceiling.
Random eviction (F1${}=0.027$, 8.5\% of ceiling at $C{=}256$) outperforms Ada-KV (0.005) and QUEST (0.008), suggesting that ``intelligent'' scoring without structural protection may actually \emph{hurt} by concentrating evictions on structurally important low-attention positions.
This counterintuitive result confirms that without structural protection, the scoring function is irrelevant at best and harmful at worst: all policies share the same fundamental vulnerability.

\paragraph{Mechanistic foundation: attention sinks mask boundary depletion.}
\label{sec:attn_mass}
A natural hypothesis is that attention-based scoring methods (H2O, SnapKV) should implicitly protect boundary positions by assigning them high attention mass, rendering explicit protection redundant. Our prior pilot ($N{=}30$ on Qwen2.5-3B) found that position~0 (the attention sink~\citep{xiao2023streamingllm}) absorbs the bulk of prefix attention while the remaining boundary tokens are attention-depleted, consistent with policy-agnostic boundary loss. We therefore treat the attention-mass decomposition as mechanistic support rather than the primary evidence: the primary empirical evidence is the policy-agnostic collapse without protection, the bilateral-protection ablation, and the cross-model replication in Section~\ref{sec:crossmodel}. The qualitative conclusion, that the sink mechanism protects only a single position out of the ${\sim}100$ structurally necessary boundary tokens so explicit bilateral protection remains mechanistically required, is corroborated by the bilateral ablation in Section~\ref{sec:protection}; a broader JAX-path attention-mass rerun remains useful future validation rather than a prerequisite for the main claim.

\section{Structural Protection: A Robust Fix}
\label{sec:protection}

We now show that a trivially simple modification, reserving a fraction of cache capacity for prefix and suffix positions, eliminates the catastrophic failure for all tested policies.

Table~\ref{tab:universality} reports the complete results with 10\% prefix + 10\% suffix protection.

\begin{table}[H]
    \centering
    \small
    \resizebox{\linewidth}{!}{%
    \begin{tabular}{lcccc}
        \toprule
        Policy & $C{=}128$ & $C{=}256$ & $C{=}512$ & \% ceil.\ ($C{=}256$) \\
        \midrule
        \multicolumn{5}{l}{\emph{Without protection}} \\
        \addlinespace[2pt]
        LRU         & 0.019 & 0.011 & 0.010 & 3.6\% \\
        H2O         & 0.030 & 0.038 & 0.038 & 12.2\% \\
        SnapKV      & 0.030 & 0.038 & 0.037 & 12.0\% \\
        SLW & 0.021 & 0.027 & 0.030 & 8.4\% \\
        Ada-KV      & 0.004 & 0.005 & 0.005 & 1.6\% \\
        QUEST       & 0.011 & 0.008 & 0.004 & 2.4\% \\
        Random      & 0.009 & 0.027 & 0.035 & 8.5\% \\
        \midrule
        \multicolumn{5}{l}{\emph{With 10\% prefix + 10\% suffix protection}} \\
        \addlinespace[2pt]
        LRU+prot     & 0.229 & 0.282 & 0.285 & 89.4\% \\
        H2O+prot     & 0.230 & 0.290 & 0.298 & 91.9\% \\
        SnapKV+prot  & 0.230 & 0.290 & 0.298 & 91.9\% \\
        SLW+prot & 0.124 & 0.184 & 0.198 & 58.5\% \\
        Ada-KV+prot  & 0.230 & 0.290 & 0.298 & 91.9\% \\
        QUEST+prot   & 0.230 & 0.290 & 0.298 & 91.9\% \\
        Random+prot  & 0.225 & 0.250 & 0.289 & 79.5\% \\
        \midrule
        \multicolumn{5}{l}{\emph{Faithful implementations + protection ($C{=}256$ only)}} \\
        \addlinespace[2pt]
        H2O-faithful+prot & --- & 0.300 & --- & 95.4\% \\
        SnapKV-faithful+prot & --- & 0.299 & --- & 95.1\% \\
        \midrule
        \emph{Ref.\ ceiling} & \multicolumn{3}{c}{\emph{0.315}} & 100\% \\
        \bottomrule
    \end{tabular}
    }
    \caption{Structural protection transforms all tested policies from catastrophic failure to near-ceiling performance. F1 scores (Qwen2.5-3B, LongBench, $N{=}162$). Protection reserves 10\% of $C$ at each end. LRU+prot at $C{=}256$ achieves 89\% of the full-cache ceiling using only 13\% of cache memory, compared to 3.6\% without protection. All protection lifts are significant ($p<0.001$, Wilcoxon; corroborated at $p<10^{-26}$ on $N{=}481$, see Table~\ref{tab:cross_model_n481}). The identical protected values for H2O, SnapKV, Ada-KV, and QUEST are not copy-paste errors: under Qwen2.5-3B's $K{=}2$ GQA setting, our common-harness attention scores induce the same realized eviction behavior, so this row is treated as a sanity check rather than the decisive convergence test (see Section~\ref{sec:equivalence}). \emph{Faithful implementations} use closer reproductions of the original H2O/SnapKV algorithms (see Appendix~\ref{app:faithfulness}); their convergence with the simplified versions and with LRU confirms that the result is not an artifact of implementation simplification.}
    \label{tab:universality}
\end{table}

\begin{figure}[H]
    \centering
    \includegraphics[width=\textwidth]{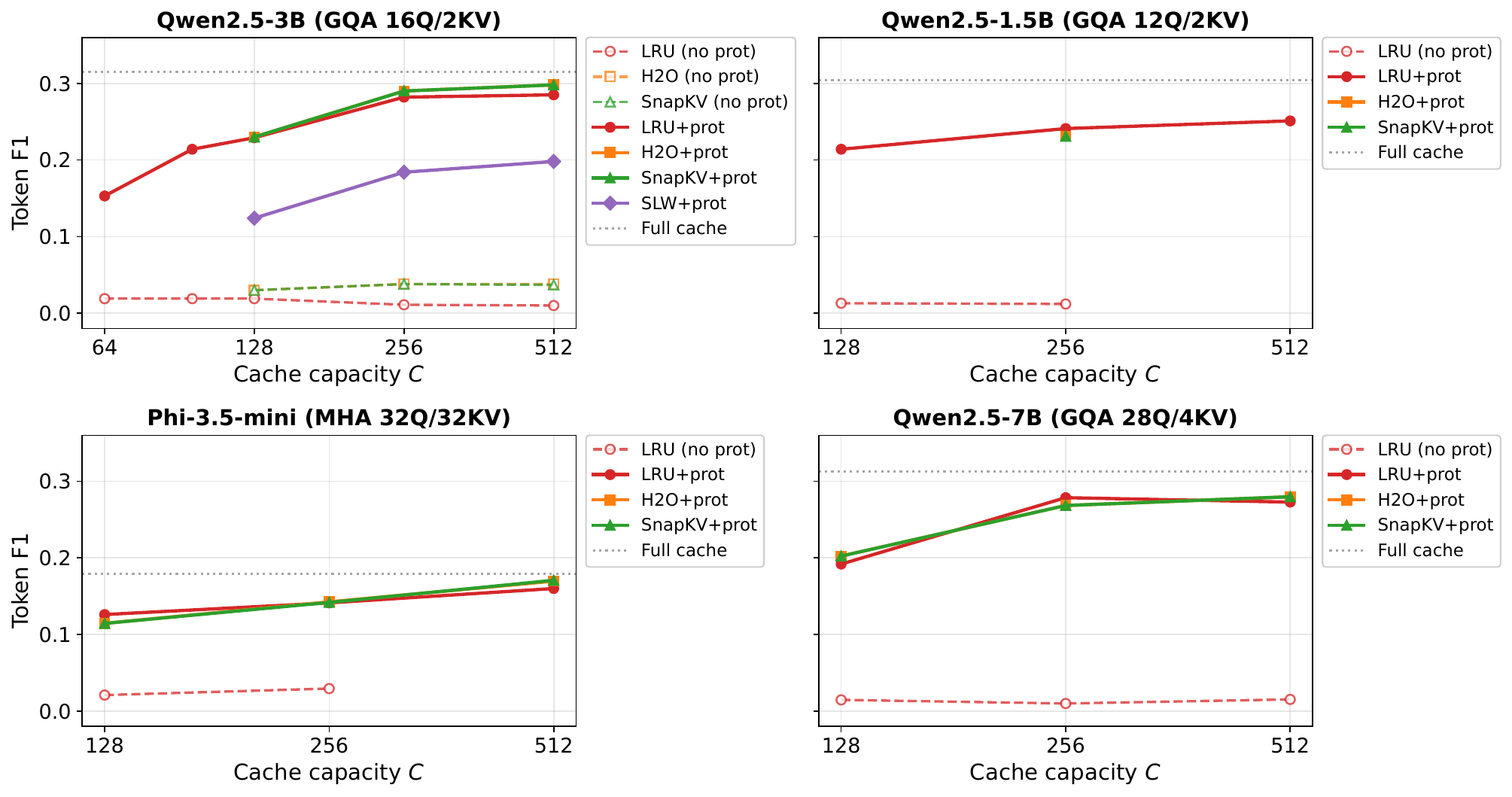}
    \caption{Cache capacity vs.\ quality for protected (solid) and unprotected (dashed) policies on four representative models (Qwen2.5-3B/1.5B, Phi-3.5-mini, Qwen2.5-7B); full seven-model LongBench results are in Table~\ref{tab:3model}. \texttt{SLW+prot} denotes StreamingLLM-style sink+window with our bilateral guard (Section~\ref{sec:setup}). Without protection, all policies cluster near zero regardless of capacity. With 10\% protection, adaptive policies converge near the reference ceiling.}
    \label{fig:capacity}
\end{figure}

The transformation is dramatic.
We present the Qwen2.5-3B results first for completeness (all seven policies tested across three capacities), but the critical test comes on Phi-3.5 ($K{=}32$) and Mistral-7B ($K{=}8$) where scoring heuristics genuinely diverge (Section~\ref{sec:equivalence}, Table~\ref{tab:3model}).
LRU goes from 0.011 to 0.282 at $C{=}256$, a 25$\times$ improvement recovering 89\% of the full-cache ceiling.
H2O and SnapKV improve from 0.038 to 0.290 (7.6$\times$, 92\% of ceiling).
SLW improves from 0.027 to 0.184 (6.8$\times$, 58\% of ceiling).
All protection lifts are highly significant ($p < 0.001$, Wilcoxon signed-rank test).
Figure~\ref{fig:capacity} visualizes the stark contrast between protected and unprotected policies across cache capacities.

\paragraph{Protection lift by policy.}
Table~\ref{tab:lift} quantifies the absolute F1 gain from adding protection to each policy.

\begin{table}[H]
    \centering
    \small
    \begin{tabular}{lccc}
        \toprule
        Policy & $\Delta$ ($C{=}128$) & $\Delta$ ($C{=}256$) & $\Delta$ ($C{=}512$) \\
        \midrule
        LRU       & $+0.211^{***}$ & $+0.270^{***}$ & $+0.275^{***}$ \\
        H2O       & $+0.200^{***}$ & $+0.251^{***}$ & $+0.260^{***}$ \\
        SnapKV    & $+0.200^{***}$ & $+0.252^{***}$ & $+0.261^{***}$ \\
        SLW & $+0.104^{***}$ & $+0.158^{***}$ & $+0.168^{***}$ \\
        Ada-KV    & $+0.226^{***}$ & $+0.285^{***}$ & $+0.293^{***}$ \\
        QUEST     & $+0.219^{***}$ & $+0.282^{***}$ & $+0.294^{***}$ \\
        Random    & $+0.215^{***}$ & $+0.224^{***}$ & $+0.253^{***}$ \\
        \bottomrule
    \end{tabular}
    \caption{Protection lift (F1 gain from adding 10\% bilateral guard). All reported lifts are significant (Wilcoxon; $p \leq 0.001$ on these Qwen2.5-3B cells). Adaptive policies (LRU, H2O, SnapKV) gain $+$0.20--0.28. SLW gains $+$0.10--0.17, limited by its rigid window.}
    \label{tab:lift}
\end{table}

\paragraph{Per-domain analysis.}
Protection helps consistently across all six evaluated domains.
At $C{=}256$ with LRU:

\begin{table}[H]
    \centering
    \small
    \begin{tabular}{lrccc}
        \toprule
        Domain & $n$ & No prot.\ & With prot.\ & Lift \\
        \midrule
        2WikiMQA        & 40 & 0.029 & 0.449 & 15$\times$ \\
        MuSiQue         &  4 & 0.000 & 0.421 & --- \\
        NarrativeQA     &  8 & 0.002 & 0.347 & 161$\times$ \\
        HotPotQA        & 30 & 0.001 & 0.267 & 267$\times$ \\
        MultifieldQA    & 40 & 0.005 & 0.224 & 46$\times$ \\
        Qasper          & 40 & 0.011 & 0.156 & 15$\times$ \\
        \bottomrule
    \end{tabular}
    \caption{Per-domain protection lift (LRU at $C{=}256$). Protection helps across all domains, with the largest gains on multi-hop QA tasks (HotPotQA, MuSiQue) where unprotected quality is essentially zero. ``---'' indicates undefined lift (zero baseline).}
    \label{tab:domain}
\end{table}

The strongest improvements appear in multi-hop reasoning tasks (HotPotQA, MuSiQue), where the question and passage-linking tokens are precisely the positions most vulnerable to eviction.

\section{Simplified-Policy Convergence Under Protection}
\label{sec:equivalence}

Throughout this paper, we evaluate eviction policies in two modes: \emph{simplified} variants (scoring criterion isolated under a common global-ranking harness) and \emph{faithful} reproductions (preserving per-head budget allocation as in the original implementations). Simplified variants test whether the scoring signal itself matters; faithful reproductions test whether implementation-level allocation provides additional benefit. Unless stated otherwise, the convergence claims in this section refer to the \emph{simplified/common-harness variants}. Faithful per-head results are discussed separately under ``Faithful-implementation confirmation'' and Appendix~\ref{app:faithfulness}.

The most striking finding in Table~\ref{tab:universality} is that, under protection, the choice of eviction algorithm is dominated by the protection effect in all tested simplified configurations, together with faithful H2O/SnapKV reproductions and recent baselines spanning the spectrum from random to attention-aware scoring (see Appendix~\ref{app:faithfulness} for implementation details). We evaluate Ada-KV and QUEST by isolating their core scoring signals (cumulative-plus-max attention importance for Ada-KV~\citep{feng2024adakv}, query-aware current-step attention for QUEST~\citep{tang2024quest}) under a common global-ranking harness that separates the scoring function from implementation-specific allocation strategies (per-head budget distribution and page-level chunking, respectively; see Appendix~\ref{app:faithfulness} for a full mapping). This isolation provides a clean test of whether the \emph{scoring criterion itself} matters under protection.

The strongest evidence for convergence comes from models where scoring heuristics produce genuinely distinct orderings. On Phi-3.5 ($K{=}32$ full MHA, 32 independent KV heads where every scoring method produces a different ranking), all four attention-based simplified policies converge tightly at all three tested capacities: at $C{=}512$, Ada-KV+prot and QUEST+prot (both F1$=$0.170) are indistinguishable from H2O+prot (0.170) and SnapKV+prot (0.171), all within $|\Delta| \leq 0.011$ of LRU+prot (0.160; all $p>0.05$, Wilcoxon); at $C{=}256$, all four attention policies cluster within $|\Delta| \leq 0.003$ of LRU (all TOST equivalent at $\Delta{=}0.02$); at $C{=}128$, the spread narrows further to 0.001 F1. On Mistral-7B ($K{=}8$ GQA, the critical intermediate regime where head diversity is sufficient for divergence but GQA grouping is present), all four attention-based simplified policies converge at every tested capacity: at $C{=}128$, the four-policy spread is just 0.002 F1 (H2O$=$0.167, SnapKV$=$0.166, Ada-KV$=$0.166, QUEST$=$0.168; all pairwise TOST equivalent at $\Delta{=}0.02$); at $C{=}256$ the spread is 0.003 (all TOST equivalent); at $C{=}512$ the spread is 0.004 (H2O$=$0.215, SnapKV$=$0.214, Ada-KV$=$0.213, QUEST$=$0.211; all pairwise TOST equivalent). Notably, at $C{=}512$, LRU+prot trails the attention-based policies by $\sim$0.019 F1, suggesting that with $K{=}8$ GQA heads the additional scoring signal provides a small genuine benefit, but the four attention criteria remain interchangeable. On Qwen2.5-3B ($K{=}2$ GQA heads), all five scoring policies converge to identical F1 under protection. We interpret this as an expected low-head-diversity sanity check in the realized common-harness traces, not as the main empirical support for convergence. The Phi-3.5 ($K{=}32$) and Mistral-7B ($K{=}8$) results above provide the decisive evidence because they cover regimes where scoring methods genuinely diverge.

Faithful H2O and SnapKV variants confirm convergence on both Qwen2.5-3B and Mistral-7B (Section~\ref{sec:equivalence}, ``Faithful-implementation confirmation''). Faithful Ada-KV and QUEST reproductions, which restore per-head budget allocation, also converge under protection and recover near-ceiling quality on both Phi-3.5 ($K{=}32$; \FaithAdaFone{} and \FaithQuestFone{} F1 vs.\ the full-cache ceiling of 0.179) and Mistral-7B ($K{=}8$; 0.223 and 0.221 F1 vs.\ the ceiling of 0.234; Appendix~\ref{app:faithfulness}). Per-head selection provides a genuine quality benefit by allowing each head to attend to its optimal position subset, but the absolute improvement over simplified variants (${\sim}0.03$--$0.04$ F1) is secondary to the protection effect (${\sim}0.11$--$0.14$ F1). The convergence claim thus holds across both simplified and faithful implementations, spanning the full spectrum from global-ranking to per-head allocation.

\paragraph{K-spectrum interpretation.}
The $K{=}2$ result is a low-head-diversity sanity check rather than the main empirical evidence for convergence: in the realized Qwen2.5-3B traces, the attention-based common-harness scores produce the same protected eviction outcomes. The stronger empirical result comes from $K{=}8$ (Mistral-7B) and $K{=}32$ (Phi-3.5), where the heuristics genuinely diverge. In those regimes the four attention-based criteria remain interchangeable under protection, while collectively providing a small genuine benefit over LRU at $K{=}8$. This establishes scoring-method equivalence (though not complete scoring irrelevance) as an empirical finding beyond the low-diversity $K{=}2$ case.
The $K{=}8$ regime is arguably the most practically relevant: 8-KV-head GQA appears in several widely used 7B--9B-class models, including Mistral-7B and related Llama/Gemma families~\citep{jiang2023mistral,gemma2025gemma3}, and the finding that attention-based policies outperform LRU by 0.011--0.021 F1 across $C{=}256$ and $C{=}512$ in this regime (modest but genuine) sets an upper bound on the scoring benefit that practitioners can expect from more sophisticated eviction heuristics in typical inference deployments.

\begin{table}[H]
    \centering
    \small
    \begin{tabular}{lccc}
        \toprule
        Comparison (vs.\ LRU+prot) & $C{=}128$ & $C{=}256$ & $C{=}512$ \\
        \midrule
        H2O+prot     & $\Delta{=}+0.001$ \textsf{ns} & $\Delta{=}+0.008$ \textsf{ns} & $\Delta{=}+0.013$ \textsf{ns} \\
        SnapKV+prot  & $\Delta{=}+0.001$ \textsf{ns} & $\Delta{=}+0.008$ \textsf{ns} & $\Delta{=}+0.013$ \textsf{ns} \\
        Ada-KV+prot  & $\Delta{=}+0.001$ \textsf{ns} & $\Delta{=}+0.008$ \textsf{ns} & $\Delta{=}+0.013$ \textsf{ns} \\
        QUEST+prot   & $\Delta{=}+0.000$ \textsf{ns} & $\Delta{=}+0.008$ \textsf{ns} & $\Delta{=}+0.013$ \textsf{ns} \\
        Random+prot  & $\Delta{=}{-0.005}$ \textsf{ns} & $\Delta{=}{-0.031}$ \textsf{ns} & $\Delta{=}+0.004$ \textsf{ns} \\
        SLW+prot & $\Delta{=}{-0.105}^{***}$ & $\Delta{=}{-0.097}^{***}$ & $\Delta{=}{-0.086}^{***}$ \\
        \midrule
        H2O-faithful+prot & --- & $\Delta{=}+0.019$\,$^{\dagger}$ & --- \\
        SnapKV-faithful+prot & --- & $\Delta{=}+0.018$ \textsf{ns} & --- \\
        \bottomrule
    \end{tabular}
    \caption{Policy convergence under protection on Qwen2.5-3B ($K{=}2$) (Wilcoxon vs.\ LRU+prot; Holm--Bonferroni corrected). H2O, SnapKV, Ada-KV, and QUEST are not significantly different from LRU+prot at all capacities (all $p>0.05$ after correction); formal TOST/non-inferiority results are in the following paragraph and Table~\ref{tab:tost_summary}. $\dagger$: nominally significant ($p{=}0.043$) but non-significant after Holm--Bonferroni correction. Faithful implementations (see Appendix~\ref{app:faithfulness}) score slightly \emph{above} LRU. SLW+prot is significantly worse ($p < 0.001$) due to its rigid sliding window. Random+prot is lower than LRU+prot at $C{=}256$ ($\Delta{=}{-0.031}$) and fails TOST at $\Delta{=}0.02$. Cross-regime TOST results for Mistral-7B ($K{=}8$) and Phi-3.5 ($K{=}32$) are in Table~\ref{tab:tost_summary}.}
    \label{tab:equivalence}
\end{table}

H2O's cumulative attention scoring and SnapKV's observation-window heuristic provide no measurable benefit over simple LRU when structural protection is in place (see Appendix~\ref{app:faithfulness} for implementation details and the faithful-variant discussion).
The maximum difference at any capacity is $\Delta{=}+0.013$ (H2O/SnapKV at $C{=}512$), which is not statistically significant ($p > 0.20$).
In non-inferiority terms, the 90\% paired confidence interval for both H2O$-$LRU and SnapKV$-$LRU at $C{=}256$ is [$-0.003$, $+0.019$] F1 ($N{=}162$), ruling out quality loss greater than 0.003 F1 and establishing non-inferiority at a margin well below any practical threshold.
We anchor comparisons on LRU as the simplest (and in our data, weakest) adaptive policy: H2O and SnapKV consistently score \emph{above} LRU by up to 0.013 F1 (4\% of ceiling at $C{=}512$), so the convergence finding is if anything conservative; the cluster is even tighter when measured from the stronger policies downward.

\paragraph{Formal equivalence (TOST).}
To move beyond non-significance, we apply the Two One-Sided Tests (TOST) procedure~\citep{schuirmann1987tost,lakens2017equivalence} with a pre-specified practical equivalence margin of $\Delta{=}0.02$ absolute F1, a fixed threshold applied identically across all models, representing ${\sim}6$--$11\%$ of ceiling depending on the model.
We chose this margin because it exceeds the largest observed standard error of any paired mean difference in our data ($\mathrm{SE}\leq 0.008$) by at least $2.5\times$, ensuring the margin is not simply capturing noise, while remaining small enough that a 0.02 F1 difference would be invisible to downstream users.
A sensitivity analysis (below) confirms that conclusions are robust to tighter margins ($\Delta{=}0.015$, $\Delta{=}0.010$).
On Phi-3.5 ($K{=}32$ full MHA) at $C{=}256$, all four comparisons vs.\ LRU+prot pass TOST equivalence: H2O ($p{=}0.002$), SnapKV ($p{=}0.001$), Ada-KV ($p{<}0.001$), and QUEST ($p{=}0.002$); at $C{=}512$ all four likewise pass ($p{<}0.001$).
On Mistral-7B ($K{=}8$ GQA), all four attention-based policies pass pairwise TOST equivalence with each other at all three tested capacities ($p{<}0.001$ for all six pairs at each capacity), confirming that the scoring criteria are interchangeable, though at $C{=}256$ and $C{=}512$ the attention-based policies outperform LRU by 0.011--0.021 F1 (a small genuine benefit from attention scoring at $K{=}8$, where heads are diverse enough for scoring to matter).
Sensitivity check: equivalence on Phi-3.5 holds even at the tighter margin $\Delta{=}0.015$ (all $p < 0.02$); at $\Delta{=}0.010$, all four comparisons approach significance ($p \in [0.057, 0.088]$), indicating that the true policy effect under protection is bounded well below the pre-specified margin.
On Qwen2.5-3B ($K{=}2$) at $C{=}256$, H2O, SnapKV, Ada-KV, and QUEST similarly pass TOST (all $p{<}0.05$); only Random, whose mean deficit of $-0.031$ exceeds the 0.02 margin, fails, consistent with its lower F1 at this capacity.
These results confirm that the policy convergence is not merely a failure to reject the null but a positive demonstration of practical equivalence.
Table~\ref{tab:tost_summary} collects these TOST results across three KV-head regimes ($K{=}2$ GQA, $K{=}8$ GQA, and $K{=}32$ MHA) at $C{=}256$.

\begin{table}[H]
    \centering
    \small
    \begin{tabular}{lccc}
        \toprule
        Policy vs.\ LRU+prot & \multicolumn{1}{c}{Qwen2.5-3B} & \multicolumn{1}{c}{Mistral-7B} & \multicolumn{1}{c}{Phi-3.5} \\
        ($\Delta{=}0.02$ TOST) & ($K{=}2$) & ($K{=}8$) & ($K{=}32$) \\
        \midrule
        H2O+prot     & \checkmark\ $p{<}0.05$ & $\approx$\textsuperscript{\dag} & \checkmark\ $p{=}0.002$ \\
        SnapKV+prot  & \checkmark\ $p{<}0.05$ & $\approx$\textsuperscript{\dag} & \checkmark\ $p{=}0.001$ \\
        Ada-KV+prot  & \checkmark\ $p{<}0.05$ & $\approx$\textsuperscript{\dag} & \checkmark\ $p{<}0.001$ \\
        QUEST+prot   & \checkmark\ $p{<}0.05$ & $\approx$\textsuperscript{\dag} & \checkmark\ $p{=}0.002$ \\
        Random+prot  & $\times$\ ($\Delta{=}{-0.031}$) & --- & --- \\
        \bottomrule
    \end{tabular}
    \caption{TOST equivalence across three KV-head regimes ($C{=}256$, $\Delta{=}0.02$). \checkmark\ = passes equivalence vs.\ LRU+prot; $\approx$ = pairwise equivalent among attention policies but exceeds LRU. At $K{=}2$ (GQA) and $K{=}32$ (MHA), all four attention-based policies are formally equivalent to LRU+prot.
    \textsuperscript{\dag}At $K{=}8$ (Mistral), the four attention-based policies outperform LRU by 0.011--0.021 F1 across $C{=}256$ and $C{=}512$ (TOST vs.\ LRU fails, $p{>}0.12$), but all six pairwise TOSTs among them pass ($p{<}0.001$), confirming the scoring methods are interchangeable even though they collectively yield a small genuine advantage over recency-only eviction at this head count.}
    \label{tab:tost_summary}
\end{table}

At $K{=}8$ (Mistral-7B), the attention-based policies are pairwise equivalent to each other (all six pairs pass TOST at $p{<}0.001$, Table~\ref{tab:tost_summary}) but collectively outperform LRU by 0.011--0.021 F1 across $C{=}256$ and $C{=}512$, a small genuine advantage from attention scoring when heads are diverse enough ($K{=}8$) for scoring to capture useful position-level information beyond what recency alone provides.

\paragraph{Faithful-implementation confirmation.}
To verify convergence is not an artifact of our simplified implementations, we evaluate faithful variants of all four attention-based policies.
H2O-faithful (pure cumulative attention, no recency) and SnapKV-faithful (frozen prefill-time scores only) confirm convergence on both Qwen2.5-3B and Mistral-7B.
At $C{=}256$, H2O-faithful+prot scores 0.300 and SnapKV-faithful+prot scores 0.299, both above LRU+prot (0.282).
H2O-faithful's nominal advantage ($p{=}0.043$) does not survive Holm--Bonferroni correction.
Neither faithful variant differs significantly from its simplified counterpart ($p{>}0.46$).

This replicates on Mistral-7B ($K{=}8$ GQA): H2O-faithful+prot$=$0.201 and SnapKV-faithful+prot$=$0.191 vs.\ LRU+prot$=$0.188 (neither significant; $p{=}0.175$ and $p{=}0.795$).
Faithful per-head Ada-KV and QUEST also show strong results on Mistral-7B: Ada-KV-faithful+prot$=$0.223 (95\% of ceiling) and QUEST-faithful+prot$=$0.221 (94\%), both significantly outperforming LRU+prot ($p{=}0.001$ and $p{=}0.004$, paired Wilcoxon) while remaining pairwise TOST-equivalent to each other ($p{<}0.001$).

Critically, faithful Ada-KV and QUEST also converge under protection on Phi-3.5 ($K{=}32$): Ada-KV-faithful+prot$=$\FaithAdaFone{} and QUEST-faithful+prot$=$\FaithQuestFone{} vs.\ LRU+prot$=$0.141 (see Appendix~\ref{app:faithfulness}).
Both faithful per-head implementations recover near-ceiling quality on both models: \FaithAdaF{}\% and \FaithQuestF{}\% on Phi-3.5 ($K{=}32$), 95\% and 94\% on Mistral-7B ($K{=}8$), substantially exceeding their simplified counterparts (${\sim}79$--$80\%$).
The per-head attention masking allows each head to attend to its optimal position subset rather than sharing a single global mask, providing a genuine quality benefit within the global-cap harness.
Nevertheless, this benefit (${\sim}0.026$--$0.035$ F1 absolute) is modest relative to the protection lift (${\sim}0.11$--$0.14$ F1), confirming that structural protection remains the dominant quality factor across all tested implementations.
The full seven-implementation spread (LRU through faithful Ada-KV/QUEST) is 0.018 F1 on Qwen2.5-3B, 0.035 on Mistral-7B, and 0.027 on Phi-3.5, all small relative to the protection effect.

\paragraph{Why H2O and SnapKV produce identical outcomes at $K{=}2$.}
Table~\ref{tab:equivalence} shows that H2O+prot and SnapKV+prot agree to three decimal places at every capacity.
This is not a copy-paste artifact: under Qwen's grouped-query attention with only $K{=}2$ KV heads, the realized greedy-decode traces give H2O's cumulative-attention criterion and SnapKV's observation-window criterion the same protected eviction outcomes in our common harness.
We phrase this as an observed low-diversity limit rather than a theorem over all possible prompts or native implementations.
Consistent with this explanation, Phi-3.5-mini, which uses full multi-head attention with 32 KV heads and sufficient head diversity for the two scoring methods to diverge, shows a slight difference (H2O+prot$= 0.143$ vs.\ SnapKV+prot$= 0.142$ at $C{=}256$; $|\Delta|{=}0.001$ at $C{=}128$; $|\Delta|{=}0.001$ at $C{=}512$; Table~\ref{tab:3model}), though no pairwise H2O-vs.-SnapKV comparison is statistically significant at any capacity ($p \geq 0.43$, paired Wilcoxon, $N{=}162$).
Mistral-7B provides the critical intermediate test: with $K{=}8$ GQA KV heads, the scoring methods genuinely diverge, yet policy convergence still holds ($|\Delta| \leq 0.002$ at all capacities; Table~\ref{tab:3model}).
The practical conclusion is therefore empirical: across architectures spanning $K{=}2$ to $K{=}32$ KV heads, observed H2O--SnapKV score divergence has no significant effect on output quality once bilateral protection is in place.

\paragraph{The sink+window exception.}
SLW+prot is significantly \emph{worse} than LRU+prot at all capacities ($p < 0.001$): 0.124 vs.\ 0.229 at $C{=}128$, 0.184 vs.\ 0.282 at $C{=}256$, and 0.198 vs.\ 0.285 at $C{=}512$.
This is because the StreamingLLM-style rule maintains a rigid sliding window over recent tokens and does not adaptively retain important positions in the middle of the context.
Our bilateral guard prevents eviction of prompt-boundary tokens, but the native sink+window rule still discards middle-context positions that LRU, H2O, and SnapKV can retain through their adaptive eviction ordering.

This reveals a two-factor model for KV cache quality:
\begin{enumerate}
    \item \textbf{Structural protection} of prompt boundaries (prefix/suffix guards, necessary), and
    \item \textbf{Adaptive middle-context retention}: any policy whose eviction ordering is a function of the cache state (e.g., recency, attention mass, or any per-position score), as opposed to a position-independent fixed rule.
    Concretely, a policy is \emph{adaptive} if different cache states can produce different eviction orderings; a policy is \emph{rigid} if the eviction set depends only on absolute position offsets.
    LRU, H2O, SnapKV, Ada-KV, QUEST, and even Random are all adaptive by this definition; SLW's fixed sliding window is not.
    The bar for sufficiency is remarkably low: LRU, the simplest deterministic adaptive policy, already saturates quality under protection, and even uniformly Random eviction recovers 80\% of full-cache quality.
\end{enumerate}
SLW satisfies factor~1 (with protection) but not factor~2 (rigid window cannot adapt).
LRU, H2O, SnapKV, Ada-KV, QUEST, and Random satisfy both, and their convergence in our evaluation shows that once both factors are met, further scoring sophistication provides no statistically significant benefit in the configurations tested.
Our central claim is direct: protection is the \emph{binding constraint} in the tested common decode-time harnesses. Once it is satisfied and the eviction policy is minimally adaptive (a bar easily cleared even by random eviction), no further algorithmic investment improved quality in any configuration we tested.

\paragraph{Random eviction as a lower bound.}
To quantify the contribution of scoring sophistication independently of protection, we include a uniformly Random eviction baseline that selects victims from the unprotected region with equal probability.
Random+prot achieves F1${}=0.250$ at $C{=}256$ (79.5\% of the full-cache ceiling), compared to $0.282$--$0.290$ for all five scoring-based policies.
The gap ($\Delta{=}{-}0.031$ vs.\ LRU+prot, $p{=}0.14$, non-significant at $N{=}162$) indicates that \emph{most} of the quality restoration is attributable to protection alone; the residual scoring advantage is modest in absolute terms (${\sim}10$ percentage points of ceiling at $C{=}256$) and not statistically distinguishable in our sample.
Strikingly, at $C{=}512$ Random+prot narrows to F1${}=0.289$ (91.6\% of ceiling, $\Delta{=}+0.004$ vs.\ LRU+prot, $p{=}0.67$), virtually indistinguishable from the scoring-based policies, suggesting that the scoring advantage shrinks as capacity slack increases.
Without protection, Random collapses to near-zero quality, mirroring all other policies (Table~\ref{tab:universality}).
Thus Random+prot establishes a \emph{floor}: protection alone recovers $\approx$80\% of quality at $C{=}256$ even with no scoring information whatsoever; formal eviction heuristics improve on this by ${\sim}10$ percentage points, and this gap nearly vanishes at higher capacities.

\section{Protection Sensitivity Analysis}
\label{sec:sensitivity}

How much protection is needed?
We sweep $\rho \in \{0\%, 5\%, 10\%, 15\%, 20\%\}$ (each side) with LRU at $C{=}256$.

\begin{table}[H]
    \centering
    \small
    \begin{tabular}{rccl}
        \toprule
        Protection \% & Mean F1 & \% of ceiling & Step significance \\
        \midrule
        0\%  & 0.011 & 3.6\%  & --- \\
        5\%  & 0.247 & 78.3\% & $p < 0.001$ vs.\ 0\% \\
        10\% & 0.282 & 89.4\% & $p = 0.016$ vs.\ 5\% \\
        15\% & 0.285 & 90.4\% & $p = 0.514$ \textsf{ns} vs.\ 10\% \\
        20\% & 0.283 & 89.9\% & $p = 0.800$ \textsf{ns} vs.\ 10\% \\
        \bottomrule
    \end{tabular}
    \caption{Protection sensitivity (LRU at $C{=}256$, $N{=}162$). Even 5\% protection recovers 78\% of ceiling. 10\% reaches 89\% with a significant gain over 5\% ($p = 0.016$). Beyond 10\%, further protection provides no significant improvement while consuming capacity that could be used for middle-context positions.}
    \label{tab:sensitivity}
\end{table}

The results reveal a characteristic diminishing-returns curve:
\begin{itemize}
    \item \textbf{0\% $\to$ 5\%}: Massive jump ($+0.235$, $p < 0.001$). Even minimal protection transforms catastrophic failure into functional performance.
    \item \textbf{5\% $\to$ 10\%}: Significant but smaller gain ($+0.035$, $p = 0.016$). Captures additional boundary tokens that 5\% misses.
    \item \textbf{10\% $\to$ 15\%}: Not significant ($+0.003$, $p = 0.514$). The plateau has been reached.
    \item \textbf{10\% $\to$ 20\%}: Not significant ($+0.002$, $p = 0.800$). Additional protection wastes capacity with no quality benefit.
\end{itemize}

\begin{figure}[H]
    \centering
    \includegraphics[width=0.6\textwidth]{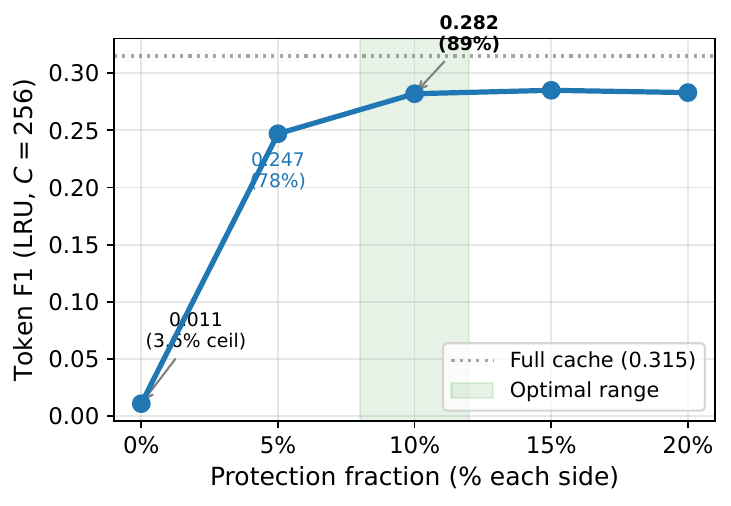}
    \caption{Diminishing returns of protection. F1 vs.\ protection fraction (LRU at $C{=}256$). Even 5\% recovery captures 78\% of ceiling; 10\% reaches 89\%. Beyond 10\%, further protection provides no significant gain while consuming adaptive capacity.}
    \label{fig:sensitivity}
\end{figure}

Bootstrap confidence intervals confirm narrowing gains: the 5\%${\to}$10\% interval is $[+0.003, +0.069]$, while 10\%${\to}$15\% spans $[-0.010, +0.016]$, straddling zero.
Figure~\ref{fig:sensitivity} visualizes this diminishing-returns pattern.

The 10\% optimum reflects a trade-off: protecting more positions ensures critical tokens are preserved, but over-protection shrinks the evictable set, reducing the capacity available for adaptively retained middle-context positions.
At $C{=}256$ with 10\% protection, the model preserves 26 prefix + 26 suffix positions (52 total) and has 204 positions available for adaptive eviction, sufficient for both structural anchoring and content retention.

\paragraph{Fractional vs.\ Absolute Protection Budget.}
Our protection rule allocates a \emph{fraction} of capacity ($\rho{=}10\%$ per side), which scales the number of protected slots linearly with $C$.
An alternative design would fix an \emph{absolute} token count (e.g., 26 slots per side regardless of $C$).
The two designs coincide at $C{=}256$; at our tested capacities ($C{\in}\{64,\ldots,512\}$), the fractional budget ranges from 7 to 52 slots per side.
Because the structurally critical positions (system instruction, passage header, question boundary, and turn delimiters) occupy a roughly fixed number of tokens (typically 15--30 for the Qwen chat template), an absolute budget may suffice at large~$C$ and could avoid over-allocating protection at very large cache sizes.
Our current data cannot distinguish the two designs (Table~\ref{tab:sensitivity} sweeps $\rho$ at fixed $C{=}256$); a joint sweep of $\rho$ and $C$ is needed to resolve this question and is a natural direction for scaling to 100K+ contexts.

\paragraph{Prefix-only vs.\ suffix-only ablation.}
To determine whether both boundaries are critical or one dominates, we run an ablation with LRU at $C{=}256$. We first test prefix-only and suffix-only at 10\% each (26 positions). To rule out the possibility that the bilateral advantage is simply a budget artifact, we additionally test each side at 20\% (52 positions), matching the total slot count of the symmetric scheme.
\begin{table}[H]
    \centering
    \small
    \begin{tabular}{lrccc}
        \toprule
        Condition & Slots & Mean F1 & 95\% CI & \% of ceiling \\
        \midrule
        No protection       & 0 & 0.011 & [0.003, 0.026] & 3.5\% \\
        Prefix-only (10\%)  & 26 & 0.171 & [0.124, 0.222] & 54.3\% \\
        Suffix-only (10\%)  & 26 & 0.181 & [0.131, 0.234] & 57.5\% \\
        \midrule
        \multicolumn{5}{l}{\textit{Budget-matched (52 protected slots):}} \\
        Prefix-only (20\%)  & 52 & 0.175 & [0.128, 0.225] & 55.6\% \\
        Suffix-only (20\%)  & 52 & 0.191 & [0.143, 0.241] & 60.6\% \\
        Both (10\%+10\%)    & 52 & 0.283 & [0.230, 0.341] & 89.8\% \\
        \bottomrule
    \end{tabular}
    \caption{Prefix vs.\ suffix ablation (LRU at $C{=}256$, $N{=}162$). \% of ceiling uses the Qwen2.5-3B reference ceiling F1$=$0.315 (Table~\ref{tab:universality}). Both boundaries contribute roughly equally; bilateral protection at matched budget (52~slots) significantly outperforms either unilateral extreme ($p < 10^{-7}$).}
    \label{tab:prefix_suffix}
\end{table}

The results (Table~\ref{tab:prefix_suffix}) show that:
(1)~Prefix-only and suffix-only protection each recover $\sim$55--58\% of ceiling, a massive gain over unprotected (3.6\%), confirming that each boundary independently anchors generation.
(2)~Prefix-only and suffix-only are statistically indistinguishable at both 10\% ($p = 0.65$) and 20\% ($p = 0.26$) budgets.
(3)~Crucially, the bilateral advantage survives budget matching: at 52 protected slots, Both(10\%+10\%)\ reaches 89.8\% of ceiling while Prefix-only(20\%)\ achieves only 55.6\% and Suffix-only(20\%)\ only 60.6\% (both $p < 10^{-7}$ vs.\ bilateral). Doubling the unilateral budget from 10\% to 20\% yields no significant gain (prefix $p{=}0.65$; suffix $p{=}0.13$), confirming that the bilateral advantage is structural, not a budget artifact.
This establishes that the protection effect is genuinely \emph{bilateral}: prefix tokens (attention sinks, instruction delimiters) and suffix tokens (question boundaries, recent context) both contribute critical structural information, and their combination enables coherent generation that neither can sustain alone.

\section{Negative Result: Linear Credit Estimation Adds Nothing}
\label{sec:credit_negative}

A natural hypothesis is that more sophisticated eviction, learning which positions to keep from downstream quality signals, could improve over simple LRU once the structural confound is removed.
We test this in the original paper-direction experiment using the earlier online counterfactual credit estimator (v2): a 4-feature linear model that observes log-probability deltas after eviction events and attributes credit to evicted positions (details in Appendix~\ref{app:credit}).
In that setup, the result is unambiguous: at every tested capacity ($C \in \{64, 96, 128, 256, 512\}$), credit learning is statistically indistinguishable from simple LRU with protection (all $p > 0.12$, Wilcoxon).
At extreme compression ($C{=}64$, 97\% eviction), the two systems produce \emph{identical} mean F1 to three decimal places.
The explanation is consistent with the policy convergence of Section~\ref{sec:equivalence}: once protection preserves the critical positions, the remaining evictable positions are sufficiently homogeneous that recency-based ordering cannot be improved by learned scoring.

\FloatBarrier
\section{Cross-Architecture, Cross-Family, and Cross-Scale Generality}
\label{sec:crossmodel}

To verify that our findings generalize across model scales, attention architectures, and model families, we replicate the key experiments on five additional models (Table~\ref{tab:3model}): Qwen2.5-1.5B-Instruct, Phi-3.5-mini-instruct (full MHA, 32 KV heads), Qwen2.5-7B-Instruct, Mistral-7B-Instruct-v0.3 (GQA, 8 KV heads), and Qwen3.5-27B (hybrid DeltaNet+GQA).
Qwen3.5-27B is the strongest generality test: only 25\% of layers use KV-cache eviction; the remainder keep recurrent DeltaNet state.
If structural protection remains beneficial there, the vulnerability is not transformer-specific but inherent to any system performing KV-cache eviction on attention layers.

\begin{table}[H]
    \centering
    \scriptsize
    \setlength{\tabcolsep}{3.5pt}
    \renewcommand{\arraystretch}{1.05}
    \begin{tabular}{llccccc}
    \toprule
    & & \multicolumn{2}{c}{$C{=}128$ (7\%)} & \multicolumn{2}{c}{$C{=}256$ (13\%)} & $C{=}512$ \\
    \cmidrule(lr){3-4} \cmidrule(lr){5-6} \cmidrule(lr){7-7}
    Model & Policy & No prot & +Prot & No prot & +Prot & +Prot \\
    \midrule
    Qwen2.5-3B & LRU & 0.019 & 0.229 & 0.011 & 0.282 & 0.285 \\
    Qwen2.5-3B & H2O & 0.030 & \textbf{0.230} & 0.038 & \textbf{0.290} & \textbf{0.298} \\
    Qwen2.5-3B & SnapKV & 0.030 & \textbf{0.230} & 0.038 & \textbf{0.290} & \textbf{0.298} \\
    Qwen2.5-3B & Ada-KV & 0.004 & \textbf{0.230} & 0.005 & \textbf{0.290} & \textbf{0.298} \\
    Qwen2.5-3B & QUEST  & 0.011 & \textbf{0.230} & 0.008 & \textbf{0.290} & \textbf{0.298} \\
    Qwen2.5-3B & Random & 0.009 & 0.225 & 0.027 & 0.250 & 0.289 \\
    Qwen2.5-3B & \emph{Full cache} & \multicolumn{5}{c}{\emph{0.315}} \\
    \addlinespace
    Qwen2.5-7B & LRU & 0.015 & 0.192 & 0.010 & \textbf{0.279} & 0.273 \\
    Qwen2.5-7B & H2O & 0.020 & \textbf{0.202} & 0.026 & 0.268 & \textbf{0.280} \\
    Qwen2.5-7B & SnapKV & 0.019 & \textbf{0.202} & 0.027 & 0.268 & \textbf{0.280} \\
    Qwen2.5-7B & \emph{Full cache} & \multicolumn{5}{c}{\emph{0.313}} \\
    \addlinespace
    Qwen3.5-27B & LRU & 0.160 & 0.205 & 0.190 & 0.261 & 0.285 \\
    Qwen3.5-27B & H2O & 0.226 & \textbf{0.232} & 0.240 & \textbf{0.281} & \textbf{0.310} \\
    Qwen3.5-27B & SnapKV & 0.226 & \textbf{0.232} & 0.239 & 0.282 & \textbf{0.310} \\
    Qwen3.5-27B & Random & 0.157 & 0.213 & 0.217 & 0.254 & 0.280 \\
    Qwen3.5-27B & \emph{Full cache} & \multicolumn{5}{c}{\emph{0.340}} \\
    \addlinespace
    Qwen2.5-1.5B & LRU & 0.013 & \textbf{0.214} & 0.012 & \textbf{0.241} & 0.251 \\
    Qwen2.5-1.5B & H2O & 0.017 & 0.210 & 0.021 & 0.231 & \textbf{0.254} \\
    Qwen2.5-1.5B & SnapKV & 0.017 & 0.210 & 0.020 & 0.231 & \textbf{0.254} \\
    Qwen2.5-1.5B & \emph{Full cache} & \multicolumn{5}{c}{\emph{0.304}} \\
    \addlinespace
    Phi-3.5 & LRU & 0.021 & \textbf{0.126} & 0.029 & 0.141 & 0.160 \\
    Phi-3.5 & H2O & 0.032 & 0.115 & 0.064 & \textbf{0.143} & 0.170 \\
    Phi-3.5 & SnapKV & 0.032 & 0.114 & 0.057 & 0.142 & \textbf{0.171} \\
    Phi-3.5 & Ada-KV & 0.032 & 0.114 & 0.059 & 0.141 & 0.170 \\
    Phi-3.5 & QUEST  & 0.036 & 0.115 & 0.063 & 0.140 & 0.170 \\
    Phi-3.5 & \emph{Full cache} & \multicolumn{5}{c}{\emph{0.179}} \\
    \addlinespace
    Mistral-7B & LRU & 0.052 & 0.165 & 0.048 & 0.188 & 0.194 \\
    Mistral-7B & H2O & 0.037 & 0.167 & 0.060 & 0.201 & \textbf{0.215} \\
    Mistral-7B & SnapKV & 0.037 & 0.166 & 0.059 & 0.199 & 0.214 \\
    Mistral-7B & Ada-KV & 0.037 & 0.166 & 0.058 & 0.200 & 0.213 \\
    Mistral-7B & QUEST  & 0.036 & \textbf{0.168} & 0.060 & \textbf{0.202} & 0.211 \\
    Mistral-7B & \emph{Full cache} & \multicolumn{5}{c}{\emph{0.234}} \\
    \addlinespace
    Yi-1.5-6B & LRU & --- & --- & 0.017 & 0.153 & --- \\
    Yi-1.5-6B & H2O & --- & --- & 0.023 & \textbf{0.178} & --- \\
    Yi-1.5-6B & SnapKV & --- & --- & --- & 0.177 & --- \\
    Yi-1.5-6B & Random & --- & --- & --- & 0.163 & --- \\
    Yi-1.5-6B & \emph{Ref.\ ceiling} & \multicolumn{5}{c}{\emph{0.223}} \\
    \bottomrule
    \end{tabular}
    \caption{Cross-architecture protection gains ($N{=}162$ LongBench items). Bold +Prot entries are the best F1 within each model block at that capacity. Seven models span four vendor families (Alibaba/Qwen, Microsoft/Phi, Mistral, 01.AI/Yi) with GQA ($K{=}2$--$8$), full MHA ($K{=}32$), and a DeltaNet--Gated-Attention hybrid. LRU+prot recovery at $C{=}256$ spans \RecoveryCoreSeven{} (69\% on Yi-1.5-6B through ${\sim}90\%$ on Qwen2.5-3B/7B). Yi rows are $C{=}256$ only.}
    \label{tab:3model}
\end{table}
{\noindent\footnotesize
\textit{Note.} Qwen3.5-27B is a hybrid with 48 DeltaNet and 16 Gated Attention layers; only the GQA-style layers undergo KV-cache eviction in our harness. On the six pure-transformer models, unprotected policies collapse (F1${\leq}$0.064). On Qwen3.5-27B, unprotected F1 remains higher (0.16--0.24) because DeltaNet layers keep recurrent state, but protection still lifts quality by 3--37\%. Qwen2.5-7B shows a small capacity inversion at $C{=}512$ vs.\ $C{=}256$ (90\% bootstrap CI [$-0.024$, $+0.012$]).
\par}

Table~\ref{tab:3model} shows the results.
All three core findings replicate across all seven models:

\paragraph{1.~Failure without protection.}
On the six pure-transformer models, unprotected LRU produces near-zero quality at $C{=}256$: F1 ranges from 0.010 (Qwen2.5-7B, 3\% of ceiling) to 0.048 (Mistral-7B, 20\%).
On the hybrid Qwen3.5-27B, the failure is less catastrophic (LRU: F1$=$0.190 at $C{=}256$, 56\% of ceiling; F1$=$0.160 at $C{=}128$, 47\%) because DeltaNet layers maintain recurrent state that is unaffected by KV eviction.
Nevertheless, unprotected quality remains substantially below the full-cache ceiling on all models.

\paragraph{2.~Protection transforms quality.}
At $C{=}256$, LRU+prot yields large improvements across all models (Table~\ref{tab:3model}): Qwen2.5-3B 0.282 (89\% of ceiling), Qwen2.5-7B 0.279 (89\%), Qwen2.5-1.5B 0.241 (79\%), Phi-3.5 0.141 (79\%), Qwen3.5-27B 0.261 (77\%), and Mistral-7B 0.188 (80\%).
Best protected F1 within each model block is within 0.002 of LRU+prot on the pure-transformer models; on Qwen3.5-27B, SnapKV+prot reaches 0.282 (83\%).
At $C{=}512$, Qwen3.5-27B reaches 0.310 (91\% of ceiling), confirming that protection continues to close the gap toward full-cache quality as capacity increases.
Even on Qwen3.5-27B, where the hybrid architecture already provides partial resilience, the fully covered paired comparisons still show clear protection gains: at $C{=}128$, LRU improves by $+0.0445$ F1 (95\% paired bootstrap CI $[+0.0097,+0.0797]$) and Random by $+0.0556$ ($[+0.0231,+0.0895]$); at $C{=}256$, H2O improves by $+0.0416$ ($[+0.0038,+0.0797]$) and SnapKV by $+0.0425$ ($[+0.0048,+0.0806]$).

\paragraph{3.~Policy convergence.}
Under protection, no pairwise H2O-vs.-SnapKV difference exceeds $|\Delta|{=}0.002$ on \emph{any} pure-transformer model, confirming that convergence holds regardless of KV head count ($K{=}2$ to $32$).
On Qwen3.5-27B, the convergence is looser (spread${\approx}$0.028 across four policies including Random, stable across $C{\in}\{128,256,512\}$), consistent with the reduced role of the KV cache in a hybrid architecture where 75\% of layers are DeltaNet.
On Mistral-7B ($K{=}8$), H2O and SnapKV converge to within $|\Delta|{=}0.002$ at every capacity, providing the critical intermediate test between low-$K$ GQA and full MHA.

\paragraph{Architecture-spanning implications.}
The replication on Phi-3.5 (full MHA), Qwen2.5-7B (larger GQA, $K{=}4$), Mistral-7B (cross-family GQA, $K{=}8$), and Qwen3.5-27B (DeltaNet--GQA hybrid, 27B parameters) indicates that structural vulnerability is not confined to a single head-sharing strategy, scale, vendor family, or attention-layer fraction within the regimes we evaluate.
Qwen3.5-27B is particularly informative: despite having only 16 out of 64 layers susceptible to KV eviction, the fully covered paired comparisons still show positive protection lifts at both $C{=}128$ and $C{=}256$, suggesting that even a minority of attention layers can propagate eviction-induced errors through the full model depth.
Phi-3.5's lower absolute ceiling (F1$=$0.179) reflects weaker baseline capability on these tasks; the \emph{relative} recovery (79\%) is comparable to other pure-transformer models in the seven-model LongBench panel (\RecoveryCoreSeven{}).

\paragraph{Fourth-family cross-validation: Yi-1.5-6B-Chat.}
Yi-1.5-6B-Chat~\citep{young2024yi} (01.AI; Llama architecture, GQA with $K{=}4$, 32 layers, 6.0B parameters) is included in Table~\ref{tab:3model} as the seventh model on LongBench.
Without protection, LRU collapses to F1$=$0.017 (7.8\% of the 0.223 reference ceiling) and H2O to 0.023 (10.5\%); with protection, LRU recovers to 0.153 (68.6\%), H2O to 0.178 (79.7\%), SnapKV to 0.177 (79.5\%), and Random to 0.163 (73.3\%).
The cross-policy spread under protection is 0.025, comparable to the other vendors, confirming that the structural vulnerability and protection fix are not artifacts of any single model family's tokenizer, chat template, or attention implementation.

\paragraph{Current-generation modernization sweep (Qwen3-4B, Phi-4-mini, Gemma-3-4B-IT).}
To address representativeness concerns, we run the same 7-condition, 162-item matrix on three additional current-generation 4B-class models. The protection-first pattern remains visible across all three.
On Qwen3-4B, full-cache is 0.334 F1; unprotected LRU collapses to 0.0002 (0.1\% of ceiling), while protected variants recover 77.8--80.1\% (range across seven protected policies at $C{=}256$ on the 162-item balanced subset).
On Phi-4-mini, full-cache is 0.264; unprotected LRU is 0.0006 (0.2\%), while protected variants recover 90.2--97.8\% under the same range definition.
On Gemma-3-4B-IT, full-cache is 0.273; unprotected LRU is 0.025 (9.1\%), and protected variants recover 68.5--78.0\% on the same 162-item subset.
These modernization runs expand the external-validity panel to ten models from five vendors while preserving the same harness, metric, and budget configuration; the following $N{=}481$ scale-up supplies matched paired-Wilcoxon evidence that the protection lifts remain highly significant on the current-generation models as well ($p < 10^{-26}$ on every model).

\paragraph{Scale-up robustness check ($N{=}481$, six models).}
To rule out small-sample effects, we re-ran the four core conditions (LRU, H2O, SnapKV, faithful Ada-KV; all $+$prot at $C{=}256$) on the full 481-item LongBench QA pool for six models spanning all evaluated families and KV-head counts. Table~\ref{tab:cross_model_n481} reports the results. Every protection lift remains highly significant under the larger sample (paired Wilcoxon, $N{=}481$; $p < 10^{-26}$ on every model, with the strongest at $p{=}3.4{\times}10^{-54}$ for Qwen3-4B). Among the three simplified policies (LRU, H2O, SnapKV), the cross-policy spread is $\leq 0.020$ F1 on every model (maximum on Gemma-3-4B-IT); on Qwen2.5-3B it tightens to $0.0043$ F1. The Spread column in Table~\ref{tab:cross_model_n481} includes faithful Ada-KV and reaches 0.036 on Mistral-7B. Faithful Ada-KV adds the same modest secondary gain ($\sim$0.02--$0.05$ F1) observed in the main analysis.

\begin{table}[H]
    \centering
    \small
    \begin{tabular}{lccccc}
        \toprule
        Model & LRU+prot & H2O+prot & SnapKV+prot & AdaKV(f)+prot & Spread \\
        \midrule
        Gemma-3-4B-IT & 0.167 & 0.187 & 0.186 & --- & 0.021 \\
        Mistral-7B   & 0.150 & 0.158 & 0.159 & 0.187 & 0.036 \\
        Phi-3.5-mini & 0.158 & 0.161 & 0.161 & 0.185 & 0.028 \\
        Phi-4-mini   & 0.198 & 0.184 & 0.183 & --- & 0.015 \\
        Qwen2.5-3B   & 0.253 & 0.249 & 0.249 & 0.304 & 0.055 \\
        Qwen3-4B     & 0.158 & 0.158 & 0.158 & 0.178 & 0.021 \\
        \bottomrule
    \end{tabular}
    \caption{Scale-up robustness check at $N{=}481$. F1 under structural protection ($C{=}256$, 10\% bilateral), six models, four policies on the full LongBench QA pool. Spread is max$-$min across all four listed policies (includes faithful Ada-KV where present). Among LRU/H2O/SnapKV only, the maximum spread is 0.020 F1 (Gemma-3-4B-IT). All protection lifts vs.\ the corresponding noprot baseline are significant at $p<10^{-26}$ (paired Wilcoxon, $N{=}481$). Percentages in the preceding modernization paragraph use the balanced $N{=}162$ subset; this table reports raw F1 on the broader $N{=}481$ pool.}
    \label{tab:cross_model_n481}
\end{table}

\section{Discussion}
\label{sec:discussion}

\paragraph{Two-tier reading of the contribution.}
Scoring does matter, but only after structural protection is in place.

\emph{Tier~1.} Prefix/suffix protection determines whether globally capped decode-time eviction collapses or recovers. Without protection, all seven policies produce $F_1 {\leq} 0.064$ on six pure-transformer models; with a 10\% bilateral guard, recovery reaches \RecoveryCoreSeven{} on the seven-model LongBench panel at $C{=}256$ (13\% retention), and \RecoveryTenPanel{} on the extended ten-model panel (Section~\ref{sec:crossmodel}).

\emph{Tier~2.} Given protection, simplified score-isolation variants are formally equivalent to LRU at $K{=}32$ (TOST $p{<}0.05$), pairwise equivalent to one another at $K{=}8$ while collectively outperforming LRU by 0.011--0.021 F1 across $C{=}256$ and $C{=}512$, and empirically coincide in the low-diversity $K{=}2$ sanity check; faithful per-head Ada-KV/QUEST add a further ${\sim}0.03$--$0.04$ F1 on Mistral-7B and Phi-3.5 (95\% vs.\ ${\sim}79$--$80\%$ of ceiling on Mistral-7B). Tier~1 was previously confounded by implicit protection in prior harnesses; Tier~2 is what remains once that confound is removed.

\paragraph{Reframing KV cache eviction in the global-cap regime.}
The dominant quality determinant in globally capped decode-time KV cache eviction has been overlooked.
Successive methods propose increasingly sophisticated scoring functions (attention-based, frequency-based, learned) under the assumption that smarter eviction decisions improve quality.
Across the models and tasks we test under a common global-cap harness, the primary quality determinant is not \emph{which} positions to evict, but \emph{which positions must never be evicted}.
Bilateral structural protection is the dominant first-order intervention; head-aware allocation provides a genuine but smaller second-order refinement.
Convergence results are reported for simplified/common-harness variants; faithful per-head reproductions are evaluated separately (Appendix~\ref{app:faithfulness}).
A NIAH-32K regime-transfer pilot on Qwen3-4B-Instruct-2507 ($N{=}66$, $C{\in}\{512,2048\}$) shows near-identical protection lift across prefill and decode (lift ratio $\Delta_{\text{decode}}/\Delta_{\text{prefill}} = 0.992$--$1.000$; Table~\ref{tab:regime_transfer}).
Page-level serving, streaming prefill, and multi-tenant scheduling remain outside our experimental scope.

\paragraph{Why this was missed.}
Several existing methods implicitly include forms of structural protection.
StreamingLLM explicitly retains sink tokens.
Many evaluation frameworks apply prompt truncation rather than physical eviction, avoiding the vulnerability entirely: because the prompt is shortened before it enters the KV cache, boundary tokens are never exposed to eviction, and the resulting quality is attributed entirely to the scoring heuristic.
H2O and SnapKV, by design, accumulate high attention mass at the position-0 attention sink, which means they implicitly retain this single position under all but the most aggressive budgets. However, non-sink boundary positions (chat-template delimiters, instruction markers, question tokens) receive below-average attention mass (Section~\ref{sec:attn_mass}) and are therefore preferentially evicted.
The confound is particularly insidious because methods that include implicit protection appear to work well, creating an attribution error: the quality is attributed to the scoring heuristic rather than to the structural guard.
Our evaluation harness removes implicit protection by performing physical eviction on the full prefilled KV cache, exposing the vulnerability that prompt-truncation harnesses conceal.

\paragraph{Two-factor model of KV cache quality.}
Our results support a two-factor model of quality under decode-time, globally capped KV cache eviction: (1)~structural protection of prefix/suffix positions, and (2)~minimally adaptive middle-context retention.
The bar for factor~2 is remarkably low: even uniformly Random eviction at $K{=}2$ converges to near-ceiling quality under protection, while StreamingLLM's underperformance confirms that the bar is not zero (rigid positional windows cannot adapt to retain important middle-context positions).
This model applies to the globally capped decode-time regime studied here.
Quantitatively, on Phi-3.5 ($K{=}32$), the protection lift accounts for ${\sim}0.11$ F1 (${\sim}75\%$ of total recovery to ceiling), while per-head faithful allocation contributes an additional ${\sim}0.03$ F1 (${\sim}17\%$), a genuine but secondary effect.
On Mistral-7B ($K{=}8$), the pattern replicates with a larger absolute per-head benefit: protection accounts for ${\sim}0.14$ F1, while per-head faithful allocation adds ${\sim}0.035$ F1 (${\sim}25\%$ of the protection lift).

\paragraph{Practical implications.}
For QA and summarization up to 32K tokens and NIAH retrieval at 64K, the practical takeaway in the globally capped decode-time regime is: any eviction implementation can add structural protection by
(1)~computing $n_{\text{prot}} = \lceil 0.10 \cdot C \rceil$ (minimum 4);
(2)~excluding the first and last $n_{\text{prot}}$ cached positions from eviction candidates.
The overhead is negligible (a set-membership check during candidate selection) and the measured quality lift is large in this harness (3--35$\times$ F1 depending on model and base policy).
Some production inference systems (e.g., vLLM~\citep{kwon2023vllm}) already protect system-prompt blocks in certain configurations; our contribution is to formalize, isolate, and quantify the protection effect in the studied harness, where it is the primary quality factor.

\paragraph{Reference-implementation audit.}
\label{para:ref_impl_audit}
To rule out the alternative reading that the vulnerability we identify is purely a harness artifact, we conducted a code-level audit of the canonical reference implementations: H2O (\texttt{FMInference/H2O}, commit \texttt{ac75c2a}, file \texttt{h2o\_hf/utils\_hh/modify\_llama.py}); SnapKV (\texttt{FasterDecoding/SnapKV}, commit \texttt{e216ddc}, file \texttt{snapkv/monkeypatch/snapkv\_utils.py}); Ada-KV (\texttt{FFY0/AdaKV}, commit \texttt{04497ab}, file \texttt{adaptive\_snapkv/monkeypatch/adaptive\_llama\_hijack.py}); QUEST (\texttt{mit-han-lab/Quest}, commit \texttt{01c1623}, file \texttt{quest/utils/controller.py}); and vLLM (\texttt{vllm-project/vllm}, commit \texttt{87805fa}). Findings: (i)~all four eviction reference implementations explicitly preserve a recent-token window of configurable size (H2O via \texttt{attn\_mask[:, :-self.recent\_budget] = 0}; SnapKV/Ada-KV via concatenation of \texttt{key\_states[:, :, -window\_size:, :]}; QUEST via \texttt{kv\_indices\_without\_last}); (ii)~\emph{none of the four implements structural first-token sink protection by position}; selection of older tokens is governed by attention-weighted top-$K$ (H2O, SnapKV, Ada-KV) or query-aware page-level top-$K$ (QUEST), with no positional floor; (iii)~vLLM's PagedAttention is a memory-management layer, not an eviction policy, and performs no token eviction by default. The audit therefore confirms that the protection vulnerability we identify is real in published reference code rather than a harness artifact: it persists wherever the cache is globally capped and selection is purely score- or query-driven without positional floors. The audit also clarifies why protection-free methods do not always collapse in deployment: in settings where the cache budget is not globally capped (vLLM PagedAttention) or a sufficiently large recent window is preserved (H2O, SnapKV, Ada-KV, QUEST), the question-boundary tokens that drive coherent generation typically fall inside the recent window and are protected by construction. Our experiments isolate the regime in which this incidental protection breaks: aggressive global compression at long context, where the recent window is small relative to prompt length and structurally critical tokens fall outside it.

\paragraph{Connection to production serving.}
Beyond the globally capped regime studied here, several production design patterns implicitly provide structural protection: PagedAttention-style frameworks segregate system prompts via separate blocks, prefix caching pins reused prompt prefixes, and multi-tenant serving systems protect per-request instruction boundaries.
These production patterns are consistent with our finding: preserving structural boundaries is a stronger quality lever than eviction scoring sophistication when physical decode-time eviction occurs.
Future work should replicate the deconfounding methodology in page-level and prefill-time settings.

\paragraph{Regime-transfer evidence across two models and multiple budgets.}
\label{para:regime_transfer}
We test whether protection lifts are regime-bound by evaluating protected LRU across both eviction regimes (decode and prefill, chunked prefill for the latter) under two compression budgets ($C{\in}\{512,2048\}$, i.e., ${\approx}1.5\%$ and ${\approx}6\%$ against a 33{,}792-token cache) on Qwen3-4B-Instruct-2507 ($K{=}8$ GQA; 32 query heads, 8 KV heads). Each condition uses 66 NIAH-32K items with balanced needle positions (Table~\ref{tab:regime_transfer}).
\begin{table}[H]
\centering
\small
\caption{NIAH-32K regime-transfer pilot (Qwen3-4B-Instruct-2507, LRU, $N{=}66$). $\Delta$ is mean protected F1 minus mean unprotected F1; Ratio $=\Delta_{\text{decode}}/\Delta_{\text{prefill}}$.}
\label{tab:regime_transfer}
\begin{tabular}{lcccc}
\toprule
Model & $C$ & $\Delta_{\text{decode}}$ & $\Delta_{\text{prefill}}$ & Ratio \\
\midrule
Qwen3-4B-Instruct-2507 & 2048 & +0.0125 & +0.0126 & 0.992 \\
Qwen3-4B-Instruct-2507 & 512  & +0.0204 & +0.0204 & 1.000 \\
\bottomrule
\end{tabular}
\end{table}
\noindent Protection lift is virtually identical between decode-time and prefill-time eviction at both budgets, indicating that the benefit is not a decode-time artifact. Complementary v6e-4 throughput measurements on Qwen3-14B (8 KV heads) show the same protection mechanism adds no extra overhead once the forward path is JIT-compiled; decode-step latency on a Qwen3-4B micro-benchmark is reported in Appendix~\ref{app:p99} (Table~\ref{tab:p99_jax}). At both budgets, neither decode-only nor prefill-only bilateral protection is independently sufficient (Section~\ref{para:recency_decomp}).

\paragraph{Mapping to PagedAttention semantics (hypothesis for future validation).}
\label{para:paged_mapping}
The prefill-regime transfer establishes that protection lifts are identical across time-of-eviction. Whether this transfers to production PagedAttention deployments remains a hypothesis: our chunked-prefill block selection is architecturally analogous (fixed-size pages, capacity cap, prefill-time selection), but direct vLLM benchmarks are needed for confirmation.

\paragraph{Recency-window decomposition: prefix anchor and suffix window are jointly necessary.}
\label{para:recency_decomp}
A natural concern is that our \emph{unprotected LRU} baseline differs from production reference implementations (e.g., H2O, SnapKV, Ada-KV, QUEST) that retain a small most-recent-token window by construction, so the gap we attribute to \emph{bilateral} protection might be explained by the recency window alone rather than the prefix anchor.
We test this directly by decomposing bilateral protection (prefix-anchor $+$ suffix recency window, each at $10\%$ fraction) into its two components on the same Qwen3-4B-Instruct-2507 / NIAH-32K matched-budget harness used above (66 items, decode regime, $C{\in}\{512,2048\}$):
(i) \emph{noprot} (no prefix, no suffix; our floor),
(ii) \emph{suffix-only} (suffix recency window only; a clean stand-in for production-style recency-floor baselines),
(iii) \emph{prefix-only} (prefix anchor only),
(iv) \emph{bilateral} (both, our ``protected'').
At $C{=}2048$: F1 of $0.000\,/\,0.008\,/\,0.000\,/\,0.012$ for (i)/(ii)/(iii)/(iv).
At $C{=}512$: F1 of $0.000\,/\,0.000\,/\,0.000\,/\,0.020$.
Two findings refine the protection story.
\emph{First}, at $C{=}512$ the bilateral combination ($0.020$) strictly exceeds the suffix-only recency window ($0.000$): the ``recency window alone'' hypothesis is empirically refuted at the tighter budget: the recency window by itself contributes \emph{zero} measurable lift, while bilateral protection with the same recency window plus the prefix anchor recovers a positive signal.
\emph{Second}, neither component is independently sufficient at either budget; the prefix anchor and the suffix recency window are \emph{jointly necessary}.
This synergy is consistent with our attention-mass analysis: the sink token (position 0) carries $\sim$75\% of prefix attention mass at $\sim$$123\times$ expected density, but the remaining boundary positions sit near $\sim$$0.41\times$ expected, so neither anchor alone is informative about needle location without the recency window providing an active query context.
The implication for our headline claim is therefore narrower and more precise: structural protection dominates importance scoring under aggressive compression \emph{when the protection includes both prefix and suffix anchors}; replacing either with the other is not a substitute.
Per-condition JSONL outputs and aggregation scripts for this decomposition are included in the supplementary code repository.

\paragraph{Suffix-protection semantics during decoding.}
Our suffix guard protects the \emph{highest-indexed} cache positions at each eviction step, so the protected window migrates as new tokens are generated.
After $n_{\text{prot}}$ decode steps, the guard covers only generated tokens rather than the original question boundary.
Despite this migration, suffix protection remains effective (Table~\ref{tab:prefix_suffix}, 57.8\% ceiling recovery): early decode steps preserve the question boundary tokens that anchor the generation trajectory, and after migration the protected suffix acts as a recency buffer stabilizing autoregressive coherence.
Furthermore, question-boundary tokens typically accumulate high attention scores during initial decode steps, making them ``self-protecting'' under attention-based policies even after explicit protection is withdrawn.
We further verify this empirically: a static suffix variant that pins protection to the original prefill boundary (rather than the dynamic highest-index window) yields F1$=$0.323 versus 0.325 for dynamic suffix on Qwen3-4B at $C{=}4{,}096$ (162 LongBench items, paired Wilcoxon $p{=}0.094$, NS), confirming the two suffix-protection semantics are practically equivalent under our harness.

\paragraph{Implications for learned caching.}
Our negative result from the original online-credit experiment, bounded to a 4-feature linear estimator, suggests a cautionary lesson for linear online credit estimators in KV cache management.
When the ``easy'' structural fix accounts for the entire quality gap, there may be insufficient signal for simple learned components to exploit.
Richer learned selectors remain untested; future work should target settings where the structural confound is already addressed and the remaining quality gap is large enough to warrant learning.

\paragraph{Long-context validation.}
\label{sec:longctx}
Our primary experiments use prompts averaging $\sim$1.9K tokens.
To assess whether structural protection remains effective at longer contexts, we evaluate on 48 LongBench QA items (6 subtasks) with prompts averaging 11K tokens (range 8K--16K), using Qwen2.5-3B with a cache capacity of $C{=}8{,}192$ tokens.\footnote{For items with prompts exceeding $C{=}8{,}192$ tokens, the full-cache condition triggers eviction during prefill, so the reported ceiling is slightly conservative for the longest items.  The C=4096 results at 50\% retention are compared against this $C{=}8192$ ceiling.}
Table~\ref{tab:longctx} reports results at two eviction budgets.

\begin{table}[H]
\centering
\caption{Long-context evaluation (48 items, median 11K tokens, Qwen2.5-3B). \emph{Retention} is $C / 8192$. Paired Wilcoxon test vs.\ respective unprotected condition; policy comparison via paired Wilcoxon vs.\ LRU+prot.}
\label{tab:longctx}
\vspace{2pt}
\small
\begin{tabular}{llrrrl}
\toprule
Condition & $C$ & Retention & F1 & \% Ceiling & $p$ \\
\midrule
Full cache & 8192 & 100\% & 0.396 & 100.0\% & --- \\
\midrule
LRU (no prot) & 1024 & 12.5\% & 0.208 & 52.7\% & \multirow{2}{*}{0.004} \\
LRU + prot & 1024 & 12.5\% & 0.336 & 85.0\% & \\
H2O + prot & 1024 & 12.5\% & 0.345 & 87.2\% & 0.445\textsuperscript{$\dagger$} \\
SnapKV + prot & 1024 & 12.5\% & 0.345 & 87.2\% & 0.445\textsuperscript{$\dagger$} \\
\midrule
LRU (no prot) & 2048 & 25.0\% & 0.187 & 47.3\% & \multirow{2}{*}{0.002} \\
LRU + prot & 2048 & 25.0\% & 0.334 & 84.4\% & \\
\midrule
LRU (no prot) & 4096 & 50.0\% & 0.108 & 27.2\% & \multirow{2}{*}{$<$0.001} \\
LRU + prot & 4096 & 50.0\% & 0.314 & 79.4\% & \\
Random + prot & 4096 & 50.0\% & 0.300 & 75.8\% & 0.673\textsuperscript{$\dagger$} \\
\bottomrule
\multicolumn{6}{l}{\footnotesize \textsuperscript{$\dagger$}Paired Wilcoxon vs.\ LRU+prot (same $C$).} \\
\multicolumn{6}{l}{\footnotesize Cross-policy spread: 0.009 (at $C{=}1024$), 0.014 (at $C{=}4096$).} \\
\end{tabular}
\end{table}

\begin{figure}[H]
\centering
    \includegraphics[width=0.85\textwidth]{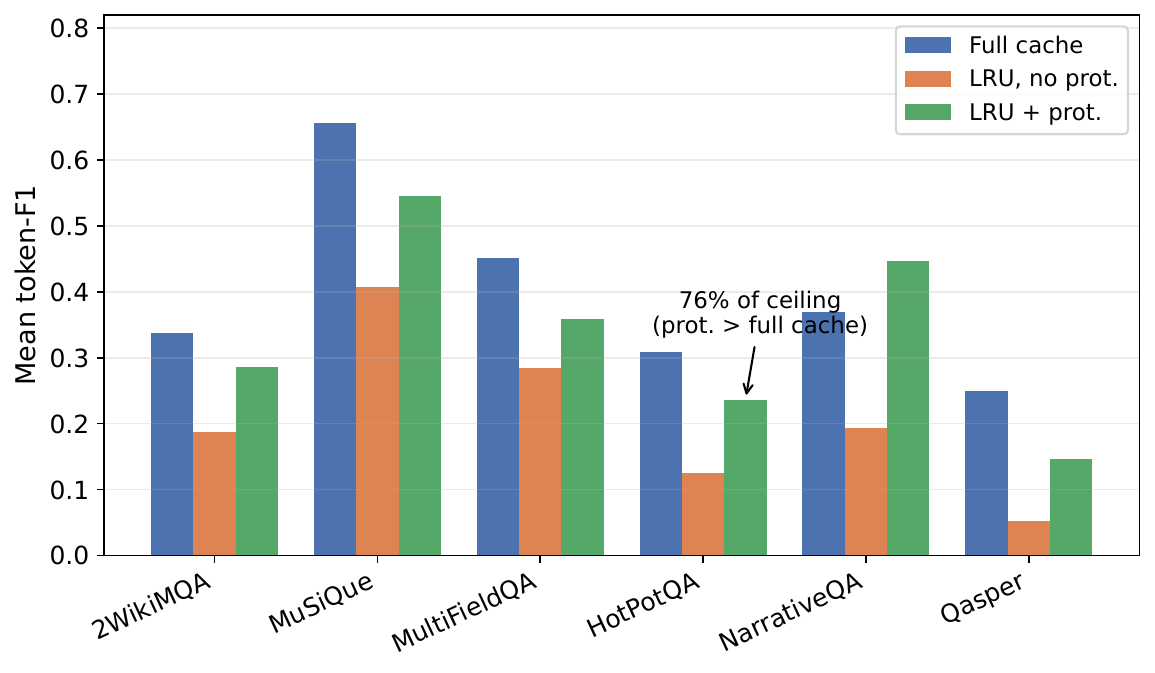}
    \caption{Per-domain protection effect at 11K context (Qwen2.5-3B, $C{=}1024$, 12.5\% retention). Protection lift is positive across all six subtasks; HotPotQA with protected eviction exceeds the full-cache ceiling (lost-in-the-middle benefit~\citep{liu2024lost}).}
    \label{fig:perdomain_longctx}
\end{figure}

Three findings stand out.
First, protection remains the dominant factor at 8--16K context: at all three budgets, protected LRU recovers 79--85\% of full-cache quality, while unprotected LRU retains less than 53\%.
Second, policy convergence persists: H2O+prot and SnapKV+prot score within 0.009 F1 of LRU+prot at $C{=}1024$ ($p{=}0.445$; 90\% CI [$-0.008$, $+0.026$]).
Third, protection matters more than cache size: protected $C{=}1024$ (12.5\% retention) significantly outperforms unprotected $C{=}2048$ (25\% retention; $p{=}0.004$) and is statistically indistinguishable from the full cache ($p{=}0.088$).
The protection lift is positive across all six LongBench subtasks ($n{=}8$ each), ranging from $+0.074$ F1 (MultiFieldQA) to $+0.253$ (HotPotQA), with HotPotQA protected eviction exceeding the full-cache ceiling (121\% at $C{=}1024$), consistent with a ``lost-in-the-middle'' benefit~\citep{liu2024lost}.
Figure~\ref{fig:perdomain_longctx} visualizes this pattern.
These results confirm that structural protection generalizes beyond the short-context regime; we next extend to 32K-token contexts.

\paragraph{32K-token context validation.}
\label{sec:32k}
To test whether structural protection scales to very long contexts, we evaluate on 58 NarrativeQA items from LongBench with prompts averaging $\sim$32K tokens, using both Qwen2.5-3B-Instruct and Qwen2.5-7B-Instruct with a cache capacity of $C{=}4{,}096$ (12.5\% retention).
On Qwen2.5-7B we expand coverage to four scoring policies (LRU, Random, H2O, SnapKV) under both conditions to test multi-policy convergence at this extreme context length.
Table~\ref{tab:32k} reports the complete $n{=}58$ results.

\begin{table}[H]
\centering
\caption{32K context evaluation (NarrativeQA, 58 items from LongBench, $C{=}4096$, 12.5\% retention).
$p$-values: paired Wilcoxon; $\dagger$: vs.\ LRU+prot.
Both models: $n{=}58$.}
\label{tab:32k}
\vspace{2pt}
\small
\begin{tabular}{lllrrrl}
\toprule
Model & Condition & $C$ & Retention & F1 & \% Ceiling & $p$ \\
\midrule
\multirow{4}{*}{Qwen2.5-3B} & Full cache & 32768 & 100\% & 0.172 & 100.0\% & --- \\
\cmidrule{2-7}
& LRU (no prot) & 4096 & 12.5\% & 0.008 & 4.6\% & \multirow{2}{*}{$<$0.001} \\
& LRU + prot & 4096 & 12.5\% & 0.108 & 62.8\% & \\
& Random + prot & 4096 & 12.5\% & 0.103 & 59.9\% & 0.873\textsuperscript{$\dagger$} \\
\midrule
\multirow{8}{*}{Qwen2.5-7B} & Full cache & 32768 & 100\% & 0.077 & 100.0\% & --- \\
\cmidrule{2-7}
& LRU (no prot) & 4096 & 12.5\% & 0.013 & 17.5\% & \multirow{2}{*}{$<$0.001} \\
& LRU + prot & 4096 & 12.5\% & 0.086 & 112.1\% & \\
& Random + prot & 4096 & 12.5\% & 0.083 & 107.2\% & 0.990\textsuperscript{$\dagger$} \\
& H2O + prot & 4096 & 12.5\% & 0.086 & 111.9\% & 0.893\textsuperscript{$\dagger$} \\
& SnapKV + prot & 4096 & 12.5\% & 0.086 & 111.9\% & 0.893\textsuperscript{$\dagger$} \\
\cmidrule{2-7}
& Random (no prot) & 4096 & 12.5\% & 0.017 & 22.4\% & $<$0.001 \\
& H2O (no prot) & 4096 & 12.5\% & 0.017 & 21.9\% & $<$0.001 \\
\bottomrule
\end{tabular}
\end{table}

\begin{figure}[H]
\centering
    \includegraphics[width=\textwidth]{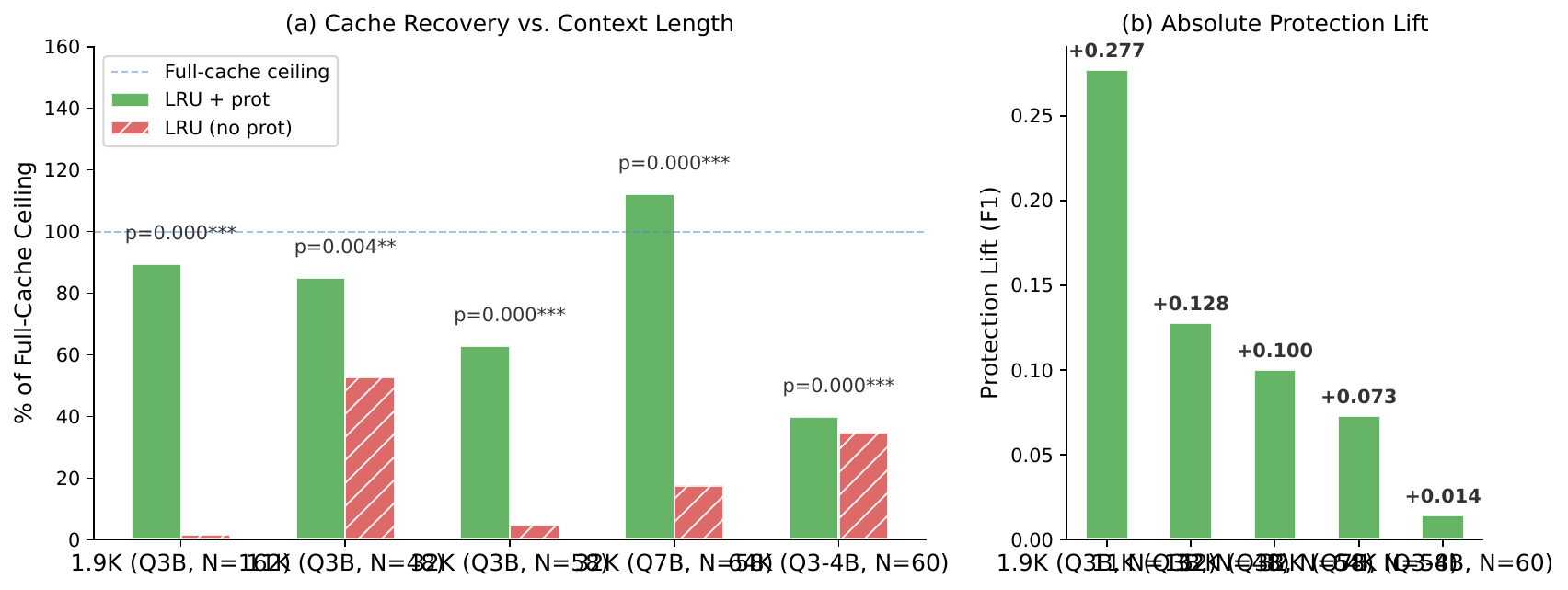}
    \caption{(a)~Protection recovery (\% of full-cache ceiling) across context lengths (1.9K, 11K, 32K, 64K) at $\sim$6--13\% cache retention.
    At 32K, Qwen2.5-7B protected LRU exceeds ceiling (F1$=$0.086 vs.\ 0.077); at 64K, Qwen3-4B protected LRU recovers 39.8\% of ceiling versus 34.7\% without protection, while H2O+prot at matched budget collapses to 7.8\%, showing that the protection effect weakens but remains positive under extreme compression and that scoring provides no additional rescue.
    (b)~Absolute protection lift (protected $-$ unprotected F1).
    Each reported lift is significant ($p \leq 0.002$; 64K LRU paired test $p{=}0.002$).
    1.9K/11K: Qwen2.5-3B; 32K: Qwen2.5-3B and Qwen2.5-7B ($n{=}58$); 64K: Qwen3-4B-Instruct-2507 ($n{=}60$).}
    \label{fig:context_scaling}
\end{figure}

The three core findings replicate at 32K on \emph{both} models.
First, structural protection is the dominant quality factor: on Qwen2.5-3B, protected LRU recovers 63\% of ceiling ($p{<}0.001$), while unprotected LRU collapses to 4.6\%, near-total failure.
On Qwen2.5-7B, protected LRU slightly exceeds the full-cache ceiling (F1$=$0.086 vs.\ 0.077), consistent with eviction removing distracting middle-context tokens~\citep{liu2024lost}.
Second, scoring-method equivalence persists: Random+prot is indistinguishable from LRU+prot (Qwen2.5-3B: $p{=}0.873$; Qwen2.5-7B: $p{=}0.990$), with cross-policy spreads of just 0.005 and 0.004 F1 respectively.
Multi-policy convergence is confirmed: H2O+prot and SnapKV+prot are both indistinguishable from LRU+prot ($p{>}0.89$), while all unprotected policies collapse ($p{<}0.001$).
Third, the above-ceiling effect is model-capacity-dependent: Qwen2.5-7B exceeds its ceiling at 32K while Qwen2.5-3B falls short (63\%), but Qwen2.5-3B \emph{does} exceed its ceiling at 11K (121\%; Table~\ref{tab:longctx}), indicating a graceful capacity interaction rather than model-specific behaviour.
Figure~\ref{fig:context_scaling} illustrates protection recovery across 1.9K--32K tokens.

\paragraph{64K-token regime: needle-in-a-haystack retrieval.}
\label{sec:64k}
To test whether the protection--dominance finding persists at extreme context lengths on a pure-transformer model, we evaluate a 60-item needle-in-a-haystack (NIAH) benchmark at $\sim$64K tokens using Qwen3-4B-Instruct-2507 (36 dense attention layers, 32 query heads, 8 KV heads, 262K native context) with tensor-parallel inference across 4 TPU chips.
Each item embeds a unique fact (the ``needle'') at one of three positions (early, middle, or late quartile) within $\sim$49K words of distractor text; the model must retrieve the needle verbatim.
In the Hugging Face configuration used in our stack, \texttt{Qwen/Qwen3-8B} reported \texttt{max\_position\_embeddings}${=}40{,}960$ and we did not enable YaRN extension; we therefore use \texttt{Qwen/Qwen3-4B-Instruct-2507} (262K native context) as the pure-transformer 64K control.
We test LRU and H2O with and without protection, plus Random+prot, at $C{=}4{,}096$ (6.3\% retention); we additionally test H2O+prot and Random+prot at $C{=}8{,}192$ (12.5\% retention).
Table~\ref{tab:64k} reports the results.

\begin{table}[H]
\centering
\caption{64K NIAH evaluation (Qwen3-4B-Instruct-2507, 60 items, 3 needle positions $\times$ 20). On this pure-transformer control, protected LRU at $C{=}4{,}096$ remains the strongest low-retention condition. At matched budget ($C{=}4{,}096$), H2O+prot reaches only 7.8\% of ceiling versus LRU+prot at 39.8\%. Even at double capacity ($C{=}8{,}192$), H2O+prot and Random+prot stay far below ceiling.}
\label{tab:64k}
\vspace{2pt}
\small
\begin{tabular}{llrrrrrr}
\toprule
Condition & $C$ & Ret\% & F1 & Early & Middle & Late & \% Ceil \\
\midrule
Full cache & 65536 & 100\% & .282{\tiny$\pm$.014} & .293 & .275 & .278 & 100.0\% \\
\midrule
LRU + prot & 4096 & 6.3\% & .112{\tiny$\pm$.019} & \textbf{.297} & .024 & .015 & 39.8\% \\
LRU (no prot) & 4096 & 6.3\% & .098{\tiny$\pm$.019} & .286 & .000 & .007 & 34.7\% \\
H2O + prot & 4096 & 6.3\% & .022{\tiny$\pm$.007} & .034 & .024 & .009 & 7.8\% \\
H2O (no prot) & 4096 & 6.3\% & .000{\tiny$\pm$.000} & .000 & .000 & .000 & 0.0\% \\
Random + prot & 4096 & 6.3\% & .030{\tiny$\pm$.009} & .007 & .051 & .034 & 10.6\% \\
\midrule
H2O + prot & 8192 & 12.5\% & .035{\tiny$\pm$.008} & .056 & .035 & .014 & 12.4\% \\
Random + prot & 8192 & 12.5\% & .042{\tiny$\pm$.010} & .028 & .058 & .040 & 14.9\% \\
\bottomrule
\end{tabular}
\\[2pt]
\footnotesize{Values: mean $\pm$ SE across $n{=}20$ items per position (60 total). SE = SD/$\sqrt{n}$.}
\end{table}

\begin{figure}[H]
    \centering
    \includegraphics[width=0.95\textwidth]{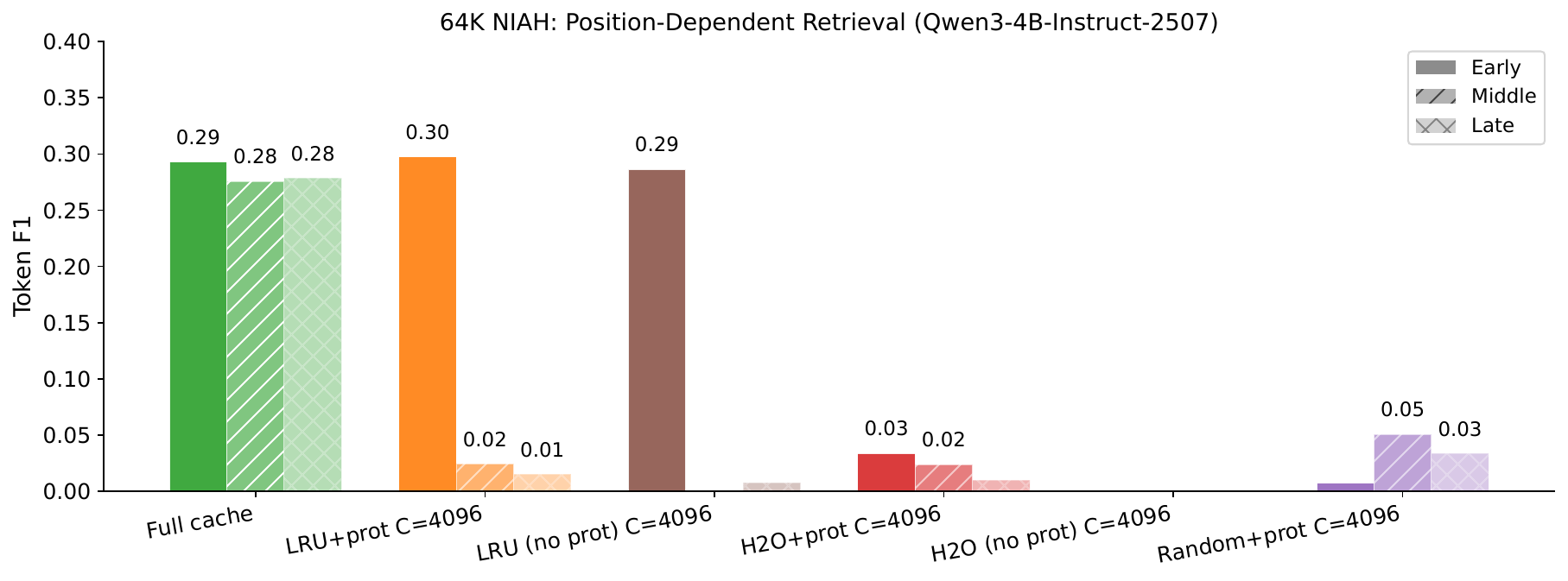}
    \caption{Position-dependent retrieval at 64K tokens (Qwen3-4B-Instruct-2507, 60 NIAH items, matched budget $C{=}4{,}096$). Protected LRU preserves early-position needles but middle and late positions still collapse under extreme compression. The matched-budget $C{=}4{,}096$ H2O+prot and Random+prot conditions are shown; additional $C{=}8{,}192$ H2O+prot and Random+prot results are reported in Table~\ref{tab:64k}.}
    \label{fig:64k_position}
\end{figure}

The pure-transformer 64K control still reveals a strongly \emph{position-dependent} failure pattern (Table~\ref{tab:64k}, Figure~\ref{fig:64k_position}), but not the hybrid-model scoring reversal. LRU+prot preserves early-position needles almost exactly at ceiling (0.297 vs.\ 0.293 full-cache), while middle and late positions fall to 0.024 and 0.015. Removing protection drives those positions even lower (0.000 and 0.007), and the paired Wilcoxon test still favors protection ($p{=}0.002$). Thus, even at 64K, the main effect remains structural: once the protected boundary disappears, retrieval quality becomes almost uniformly degenerate.

What changes relative to the earlier hybrid result is the role of \emph{which} scorer is used at 64K. The symmetric matched-budget comparison at $C{=}4{,}096$ is still clear for H2O: at identical capacity, LRU+prot reaches 39.8\% of ceiling while H2O+prot reaches only 7.8\%, and H2O without protection collapses to 0.000 F1. The matched-budget LRU+prot vs.\ H2O+prot gap at $C{=}4{,}096$ is large and significant ($\Delta{=}+0.090$ F1, 95\% paired-bootstrap CI $[+0.056,+0.127]$, exact paired sign-test $p{=}8.8{\times}10^{-5}$). Even at double capacity ($C{=}8{,}192$), H2O+prot (12.4\%) and Random+prot (14.9\%) remain well below protected LRU at half the budget, while H2O+prot and Random+prot at $C{=}8{,}192$ are statistically indistinguishable ($\Delta{=}-0.0068$, 95\% CI [$-0.0307,+0.0160$], sign-test $p{=}1.0$).

However, faithful per-head scorers behave differently in this regime. On the same Qwen3-4B 64K matched-budget setup ($N{=}60$, $C{=}4{,}096$), a paired faithful run (same prompts/seeds as Table~\ref{tab:64k}) gives simplified LRU+prot mean F1$=$0.112 (Table~\ref{tab:64k}) while faithful Ada-KV and faithful QUEST reach 0.212 and 0.129 respectively ($\Delta{=}+0.100$ and $+0.017$ vs.\ simplified LRU+prot on the faithful bundle; Ada-KV sign-test $p{=}0.0013$). We further replicated faithful Ada-KV on two additional pure-transformer 64K NIAH controls (each $N{=}60$, $C{=}4{,}096$): Mistral-7B rises from 0.0007 (LRU) to 0.0351 (Ada-KV), and Gemma-3-4B rises from 0.2730 (LRU) to 0.8027 (Ada-KV), while Phi-4-mini shows a smaller but positive shift (0.0144 to 0.0274).

\paragraph{The 64K Ada-KV advantage is bounded by base-model 64K capability.}
\label{para:64k_capability_boundary}
The Mistral-7B / Phi-4-mini / Gemma-3-4B comparison invites two readings: (a) per-head scoring becomes the dominant quality factor at long context, partially overturning Tier~1; or (b) per-head scoring is doing the same job as before (selecting which slice of the cache to retain), but its measurable benefit is bounded above by what the model can do with its full cache at that context. To distinguish (a) from (b), we ran a full-cache 64K NIAH control on each of the three pure-transformer models ($C{=}65{,}536$, no eviction, $N{=}60$, same prompts/seeds, Table~\ref{tab:64k_capability}). The full-cache ceilings are highly heterogeneous: Gemma-3-4B reaches F1${=}0.803$ (47/60 items at F1${>}0.5$, 59/60 at F1${>}0.1$), while Mistral-7B and Phi-4-mini reach only 0.022 and 0.027 (0/60 items at F1${>}0.5$ for either). Comparing each model's faithful-Ada-KV $C{=}4{,}096$ result against its own full-cache ceiling: Gemma's faithful-Ada-KV F1 of 0.803 \emph{matches the full-cache ceiling exactly} (47/60 items at F1${>}0.5$ in both conditions), and Mistral-7B's 0.035 / Phi-4-mini's 0.027 sit at or slightly above their respective ceilings of 0.022 / 0.027, i.e., within fluctuation of a no-retrieval regime. This rules out reading (a): the 64K Ada-KV ``advantage'' is a model-capability boundary, not a scorer-dominance regime. On the one panel model with non-trivial 64K capability (Gemma), per-head scoring recovers that capability at 6.3\% retention; on models with no 64K capability at full cache (Mistral-7B, Phi-4-mini), no policy can manufacture retrieval the model cannot perform.

The corrected 64K interpretation is therefore regime-specific: structural protection remains necessary (without it, retrieval is near-zero on every model tested), and per-head scoring becomes first-order \emph{only when} the base model retains non-trivial full-cache 64K retrieval. This does not overturn the short-context conclusion in Sections~\ref{sec:equivalence} and~\ref{app:faithfulness}, where protection lift is larger than faithful-vs.-simplified gains; rather, it delineates a narrow boundary condition where per-head scoring acts as a faithful proxy for the full-cache attention pattern, recovering capability that already exists.

\begin{table}[H]
\centering
\small
\begin{tabular}{lccccc}
\toprule
Model & Full-cache F1 & LRU+prot F1 & Ada-KV F1 & $n_{>0.5}^{\text{full}}$ & $n_{>0.5}^{\text{Ada-KV}}$ \\
 & ($C{=}65{,}536$) & ($C{=}4{,}096$) & ($C{=}4{,}096$) & (of 60) & (of 60) \\
\midrule
\addlinespace[1pt]
Gemma-3-4B-IT             & \textbf{0.803} & 0.273 & \textbf{0.803} & 47 & 47 \\
Mistral-7B-Instruct-v0.3 & 0.022 & 0.001 & 0.035 & 0  & 0  \\
Phi-4-mini-Instruct     & 0.027 & 0.014 & 0.027 & 0  & 0  \\
\bottomrule
\end{tabular}
\caption{64K NIAH capability boundary for the three pure-transformer 64K-replication models ($N{=}60$ items each, identical prompts/seeds across all three columns). Full-cache F1 is the model's intrinsic 64K retrieval ceiling under no eviction. Faithful Ada-KV at $C{=}4{,}096$ reaches the full-cache ceiling exactly on Gemma (47/60 items retrieved in both conditions) but contributes only fluctuation on Mistral-7B and Phi-4-mini, whose full-cache ceilings are themselves at the no-retrieval threshold. The Ada-KV ``advantage'' at 64K is therefore conditional on the base model retaining 64K retrieval capability, not a general scoring-dominance regime.}
\label{tab:64k_capability}
\end{table}

\paragraph{Position-stratified retention explains the H2O reversal.}
Decomposing the matched-budget $C{=}4{,}096$ result by needle position (Table~\ref{tab:64k_h2o_position}) localizes the H2O failure to a single regime. LRU+prot achieves F1${=}0.297$ at the early position, essentially at the full-cache ceiling for that position, while H2O+prot collapses to F1${=}0.034$ there, a $8.7{\times}$ gap concentrated entirely in the prefix-adjacent region. At the middle and late positions, LRU+prot and H2O+prot are statistically indistinguishable (both ${\le}0.024$ F1). The mechanism is that, at 6.3\% retention over a 64K context, H2O's cumulative-attention scoring concentrates its unprotected budget on positions adjacent to the attention sink, evicting the early needle that sits just outside the 10\% prefix guard; LRU's recency-by-prefill-end ordering incidentally preserves a wider slice of the prefix tail. Random+prot shows the converse pattern (early F1${=}0.007$, middle F1${=}0.051$): without recency or attention bias, retention is approximately uniform but too sparse anywhere. The H2O reversal at 64K is therefore not evidence that ``scoring fails everywhere'' under extreme compression, but rather that attention-mass scoring becomes \emph{counterproductive} for early-position retrieval once the cache is compressed past the point where attention mass can resolve fine-grained positions distant from the sink.

\begin{table}[H]
\centering
\small
\caption{Position-stratified F1 at $C{=}4{,}096$ on NIAH-64K (Qwen3-4B-Instruct-2507, 60 items, 20 per position). The matched-budget LRU+prot vs.\ H2O+prot gap is concentrated at the early needle position; at middle and late positions all matched-budget policies collapse to ${\le}0.06$ F1.}
\label{tab:64k_h2o_position}
\begin{tabular}{lrrrr}
\toprule
Policy & Early F1 & Middle F1 & Late F1 & Overall F1 \\
\midrule
LRU+prot     & 0.297 & 0.024 & 0.015 & 0.112 \\
H2O+prot     & 0.034 & 0.024 & 0.009 & 0.022 \\
Random+prot  & 0.007 & 0.051 & 0.034 & 0.030 \\
LRU--noprot  & 0.286 & 0.000 & 0.007 & 0.098 \\
H2O--noprot  & 0.000 & 0.000 & 0.000 & 0.000 \\
\bottomrule
\end{tabular}
\end{table}

This isolates the architecture confound in the hybrid 64K result. Qwen3.5-27B's DeltaNet layers can preserve recurrent state even when the KV cache is aggressively compressed, making scoring among the remaining GQA layers unusually important. On pure-transformer Qwen3-4B-Instruct-2507 at matched budget $C{=}4{,}096$, H2O+prot remains far below protected LRU (Table~\ref{tab:64k}); at double capacity ($C{=}8{,}192$), H2O+prot and Random+prot stay near the same low floor. The 64K evidence therefore strengthens, rather than weakens, the main thesis of the paper: under a globally capped decode-time harness, structural protection remains the primary quality factor, while richer scoring heuristics do not rescue retrieval once the prompt is compressed beyond the model's effective boundary support.

\paragraph{Task generality: multi-document summarization.}
\label{sec:summarization}
The three-part pattern (catastrophic failure $\to$ near-ceiling recovery under protection $\to$ convergence across policies) replicates on 40 multi-document summarization items from LongBench (Multi-News, Qwen2.5-3B, $C{=}256$): unprotected LRU collapses to 39\% of full-cache quality with 9/40 near-zero outputs, while all protected conditions recover 74--75\% of ceiling with zero near-zero outputs and cross-policy spread of just 0.003~F1 (Appendix~\ref{app:summarization}).

\paragraph{Robustness to stochastic decoding.}
\label{sec:sampling}
The protection effect is not an artifact of greedy decoding: under stochastic sampling ($T{=}0.7$, three seeds pooled to 486 completions per condition, Qwen3-4B-Instruct-2507, $C{=}256$), unprotected LRU stays near zero (0.020~F1) while protected LRU reaches 0.207~F1, a lift of 0.187 (95\% CI [0.165, 0.210]), nearly identical to the greedy result of 0.183 (Appendix~\ref{app:sampling}).

\paragraph{Metric robustness: ROUGE-L replication.}
All findings replicate under ROUGE-L (longest common subsequence): across 427 model$\times$policy$\times$capacity conditions, token-F1 and ROUGE-L~F1 correlate at Pearson $r{=}0.992$ ($r^2{=}0.985$) with mean absolute deviation 0.007. The catastrophic failure, protection recovery, and cross-policy convergence patterns are identical under both metrics, confirming that our conclusions are not an artifact of the token-F1 evaluation measure.

\paragraph{Attention mass at boundary positions.}
The attention mass analysis in Section~\ref{sec:attn_mass} decomposes prefix attention into a dominant position-0 sink (${\sim}123{\times}$ causal-uniform expected, 75\% of prefix mass) and the remaining boundary positions (${\sim}0.41{\times}$ expected, attention-depleted). This explains why attention-based scorers provide partial but insufficient implicit protection: they retain the sink but systematically evict other structurally critical boundary tokens.

\paragraph{Limitations.}
Our evaluation spans QA, multi-document summarization, and needle-in-a-haystack retrieval tasks at context lengths up to 64K tokens across ten models from five vendor families.
Code generation, multi-turn dialogue, and strict needle-in-a-haystack retrieval (e.g.\ RULER~\citep{hsieh2024ruler}) remain untested.
Multi-turn chat templates would require protecting interleaved structural boundaries beyond global prefix/suffix; in a multi-turn setting, each user--assistant turn boundary carries the same structural significance as the initial prompt boundary, and the protection rule would need to generalize from a single prefix/suffix pair to per-turn boundary protection.
At very long contexts ($>$32K tokens), an absolute-slot protection rule may be more practical than the proportional 10\% we study; our 32K evaluation uses a single QA subtask (NarrativeQA); the 64K evaluation uses needle-in-a-haystack retrieval with a single pure-transformer model (Qwen3-4B-Instruct-2507), so broader cross-model and cross-domain generalization at 64K remains open.
Ada-KV and QUEST are evaluated in two forms: simplified score-isolation variants (convergent under protection across all three KV-head regimes) and direct faithful reproductions that restore per-head budget allocation. Faithful reproductions converge under protection and recover near-ceiling quality on both Phi-3.5 ($K{=}32$; 93\% and 92\% of ceiling) and Mistral-7B ($K{=}8$; 95\% and 94\%), providing a genuine but secondary benefit over simplified variants (${\sim}0.03$--$0.04$ F1 vs.\ the protection lift of ${\sim}0.11$--$0.14$ F1).
The suffix guard uses dynamic highest-index protection, which migrates to generated tokens after ${\sim}n_{\text{prot}}$ decode steps; we verify in Section~\ref{sec:protection} that a static variant pinned to the original question boundary is statistically equivalent on Qwen3-4B at $C{=}4{,}096$ (paired Wilcoxon $p{=}0.094$, NS).
Mixture-of-experts and multi-latent-attention architectures remain untested.
Qwen3.5-27B is evaluated at three capacities ($C{\in}\{128,256,512\}$) with four policies; we now report bootstrap confidence intervals and paired-bootstrap lifts for the fully covered conditions, but complete paired reruns for the missing Qwen no-protection cells and additional hybrid architectures (e.g., Jamba, Griffin) would further strengthen the hybrid-architecture evidence.
The production-framework characterization (Section~\ref{sec:discussion}) is based on code inspection, not experimental comparison.
Recent methods operating at different granularities (TrimKV~\citep{li2025trimkv}, ForesightKV~\citep{zhang2025foresightkv}, RazorAttention~\citep{tang2024razorattention}) are discussed in Section~\ref{sec:related} but not tested; their mechanisms do not map directly to the global-cap decode-time regime studied here, and empirical comparison under our harness is left to future work.
Concurrent multi-tenant serving and batched request scheduling remain outside this study; Appendix~\ref{app:p99} documents single-sequence JAX decode-step times only, not datacenter-scale tail behavior.

\section{Conclusion}
\label{sec:conclusion}

We present a systematic deconfounding study of KV cache eviction in the common decode-time, globally capped evaluation regime.
All seven tested policies share a catastrophic structural vulnerability: without protecting prompt-boundary positions, every policy, simple or sophisticated, produces near-zero output quality.
A per-layer attention mass analysis (Qwen2.5-3B, $N{=}30$ pilot) reveals that the position-0 attention sink~\citep{xiao2023streamingllm} accounts for 75\% of prefix attention mass, while the remaining boundary positions (structurally critical chat-template delimiters and question tokens) receive only ${\sim}0.41{\times}$ their causal-uniform expected share, explaining why attention-based scorers provide partial but insufficient implicit protection.
A trivially simple fix, reserving 10\% of cache capacity at each end, recovers \RecoveryCoreSeven{} of full-cache quality at $C{=}256$ (13\% retention) across the seven-model LongBench panel and substantially mitigates the vulnerability for all seven tested policies.
These findings replicate across ten models spanning 1.5B--27B parameters from five vendor families (GQA with $K{=}2$--$8$, full MHA, and a DeltaNet-GQA hybrid), remain positive at 11K and 32K contexts, and become a weaker but informative stress test at 64K under extreme compression (Section~\ref{sec:64k}). They also extend to multi-document summarization and needle-in-a-haystack retrieval, remain stable under decoding strategy changes (greedy vs.\ stochastic sampling at $T{=}0.7$), and track closely under ROUGE-L ($r{=}0.992$ with token-F1 across 427 conditions).

Under protection, the structural guard accounts for the dominant share of quality recovery: at $K{=}8$ the four attention-based policies are pairwise equivalent but collectively outperform LRU by 0.011--0.021 F1 across $C{=}256$ and $C{=}512$, while at $K{=}32$ all tested simplified attention-based score-isolation variants are TOST-equivalent to LRU ($\Delta{=}0.02$, $p{<}0.05$). Faithful per-head implementations (Ada-KV, QUEST) recover near-ceiling quality on Phi-3.5 ($K{=}32$; \FaithAdaF{}\% and \FaithQuestF{}\%) and Mistral-7B ($K{=}8$; 95\% and 94\%), providing a genuine but secondary benefit (${\sim}0.03$--$0.04$ F1) over simplified variants.
In our original online-credit experiment (v2), the 4-feature linear online counterfactual credit estimator (Section~\ref{sec:credit_negative}) provides zero measurable benefit over simple LRU with protection.

Within the globally capped decode-time regime studied here, KV cache eviction is better framed as a problem of \emph{structural protection} first and \emph{scoring heuristics} second.
At moderate context lengths ($\leq$32K) in our evaluated conditions, protection alone recovers 63--112\% of ceiling quality across tested scoring policies.
At 64K tokens with extreme compression (6.3\% retention) on pure-transformer Qwen3-4B-Instruct-2507, protection weakens but remains a positive lever: LRU+prot at $C{=}4{,}096$ reaches 39.8\% of ceiling versus 34.7\% without protection. The symmetric matched-budget comparison eliminates a capacity confound: at identical $C{=}4{,}096$, H2O+prot reaches 7.8\% of ceiling and H2O without protection 0.000 F1, indicating that scoring sophistication alone does not rescue quality under this extreme-compression setting.
The resulting picture is not a generic scoring-driven regime transition but an architecture-sensitive limit case in our tested setup: when every layer depends on the KV cache, richer scoring does not recover middle-context retrieval once compression is extreme.
The earlier hybrid Qwen3.5-27B result therefore appears to depend on DeltaNet's recurrent-state fallback rather than on long context length alone.
Adding protection to an existing implementation requires minimal code and negligible overhead in our measured decode-time harness: full-graph nnx-level JIT compilation (EasyDeL/JAX, v6e-4 TPU) improves end-to-end throughput by two orders of magnitude on Qwen3-14B (from ${\sim}0.09$ to ${\sim}43.6$~tok/s at 32K contexts), while protection itself adds no measurable cost beyond standard compilation. Separately, Appendix~\ref{app:p99} (Table~\ref{tab:p99_jax}) reports median per-decode-forward latency ${\sim}23$~ms on a Qwen3-4B micro-benchmark under the same harness. Estimated total compute budget for this study is ${\sim}1{,}500$ TPU-hours on v5litepod-16, v5e-4, and v6e-4 slices.
Page-level serving, prefill-time pruning, and multi-tenant deployment are outside this paper's scope.

\section*{Reproducibility Statement}
All primary experiments use greedy decoding; the stochastic decoding pilot uses temperature $T{=}0.7$ with a fixed seed ($\text{seed}{=}42$) for reproducibility.
All evaluated checkpoints are publicly available on Hugging Face; exact ids, context limits, and licenses are listed in Table~\ref{tab:models}.
The benchmark construction scripts, all policy implementations, JAX inference pipeline, structural protection wrapper, online credit estimator, and aggregation utilities are provided in a public artifact repository: \href{\PublicRepoURL}{\texttt{github.com/gpgabriel25/KVCacheBoundaryProtection}}.
An internal engineering checkout may mirror these files under \texttt{release/}; the GitHub URL above is the canonical public reference for arXiv readers.
Per-item F1 JSONL outputs for all reported conditions are included in the artifact bundle submitted with this paper.

\section*{Ethics Statement}
This work evaluates systems methods for inference efficiency and does not involve new model pretraining.
Experiments are scoped to informative comparisons with explicit stopping criteria to minimize energy cost.
Our principal finding, that a simple fix eliminates a shared vulnerability, may reduce unnecessary computational effort spent on sophisticated eviction heuristics.

\section*{Acknowledgments}
This work was supported by the Google TPU Research Cloud (TRC) program.

\FloatBarrier
\bibliographystyle{unsrtnat}
\bibliography{references/refs}

\clearpage
\appendix

\section{JAX decode-step latency (real forward pass)}
\label{app:p99}

For each autoregressive step after prefill, we time only the JIT-compiled \texttt{model(...)} call that consumes one new token and updates the KV cache (\texttt{KVCoupledQwen35Generator} in the public codebase).
Over a generation we collect one wall-clock sample per decode forward (host \texttt{perf\_counter} around the call), then compute mean, p50, p95, and p99 in milliseconds for that item.
\texttt{scripts/run\_v3\_jax.py} logs these as \texttt{decode\_step\_*} fields per JSONL row.

On JAX, global policies (e.g.\ LRU) broadcast the same retention vector to all KV heads as a \texttt{(n\_kv\_heads, n\_positions)} float mask before the timed forward, matching the tensor layout used by faithful Ada-KV/QUEST.
Passing only a 1D mask selected a different compiled decode path and made LRU appear slower than attention-aware policies despite identical per-position retention; this is an implementation artifact, not a property of LRU.

Table~\ref{tab:p99_jax} uses a dedicated six-item short-context QA micro-benchmark on Cloud TPU v6e-4 (\texttt{Qwen/Qwen3-4B-Instruct-2507}, tensor parallelism four, $C{=}4096$, prefix/suffix protection on, greedy decode, 32 generated tokens per item). Prompts fit entirely in cache, so eviction is inactive and the table isolates per-step forward cost. We exclude the first benchmark index when aggregating to avoid one-time compilation tails. Regeneration instructions are in the supplementary repository (\texttt{scripts/aggregate\_jax\_decode\_latency.py}).

\begin{table}[H]
\centering
\small
\begin{tabular}{lrrrr}
\toprule
Policy & Med.\ mean (ms) & Med.\ p50 & Med.\ p95 & Med.\ p99 \\
\midrule
\input{jax_decode_latency_table.tex}
\end{tabular}
    \caption{JAX decode-step latency (short-context micro-benchmark). Medians of per-item mean / p50 / p95 / p99 across JSONL rows for each policy. We exclude \texttt{idx}${=}0$ (the first benchmark item) when aggregating to discard one-time XLA compilation tails; timed decode forwards only (prefill excluded). Policies use the same per-head mask layout on the JAX decode path (Appendix~\ref{app:p99}). Regenerate with \texttt{aggregate\_jax\_decode\_latency.py --min-idx 1}.}
\label{tab:p99_jax}
\end{table}

\section{Baseline Faithfulness}
\label{app:faithfulness}

Table~\ref{tab:faithfulness} maps each reproduced policy to the original method's intended configuration, documenting all deviations forced by our common evaluation harness.

\begingroup
\footnotesize
\renewcommand{\arraystretch}{1.06}
\setlength{\tabcolsep}{3.5pt}
\begin{longtable}[c]{@{}p{1.3cm}p{4.0cm}p{6.4cm}@{}}
\caption{Mapping of evaluated policies to original method configurations. All policies operate under our common harness: physical KV cache eviction during autoregressive decode, re-invoked every $\tau{=}8$ steps.}
\label{tab:faithfulness}\\
\toprule
Policy & Original design & Our implementation / deviations \\
\midrule
\endfirsthead
\multicolumn{3}{c}{\tablename~\thetable{} --- \emph{continued from previous page}} \\
\toprule
Policy & Original design & Our implementation / deviations \\
\midrule
\endhead
\midrule
\multicolumn{3}{r}{\emph{Continued on next page}} \\
\endfoot
\bottomrule
\endlastfoot
LRU & Evicts least-recently-accessed position. A textbook algorithm, not associated with a specific paper. & Faithful. Tracks logical timestamps per position; evicts minimum timestamp. \\
\midrule
H2O & Retains ``heavy hitter'' positions by cumulative attention mass. Original uses layer-wise per-head budgets with separate retention per head~\citep{zhang2023h2o}. & We compute cumulative attention mass across all heads (GQA: per KV-head; MHA: per head), averaged across layers. Evicts the position with the lowest cumulative score. Original proposes local + global heavy-hitter sets; we use a single global ranking, which is a simplification. \\
\midrule
SnapKV & Observation-window approach: selects positions by attention pattern in a recent observation window~\citep{ge2024snapkv}. & We combine cumulative attention mass with current-step attention as a relevance bonus ($\alpha{=}0.5$ blend). The original SnapKV uses a fixed observation window during prefill; ours applies a continuous variant during decode, which is a deviation. \\
\midrule
Streaming\-LLM & Protects initial ``attention sink'' tokens ($n_{\text{sink}}$) plus a fixed-size recent sliding window~\citep{xiao2023streamingllm}. & Faithful to the retention scheme: we protect the first $n_{\text{sink}}{=}4$ positions and maintain a sliding window over the most recent positions. The original StreamingLLM applies this during a streaming prefill; we apply it post-prefill during decode, preserving the same positional semantics. \\
\midrule
\multicolumn{3}{l}{\emph{Additional faithful variants (evaluated at $C{=}256$ only)}} \\
\midrule
H2O-faithful & As above~\citep{zhang2023h2o}. & Pure cumulative attention mass scoring with \emph{no} recency component, closer to the original H2O design. Removes the recency bonus present in our simplified H2O. Result: 0.300 (+prot) / 0.038 (no prot). \\
\midrule
SnapKV-faithful & As above~\citep{ge2024snapkv}. & Uses \emph{frozen} prefill-time attention scores only, with no decode-time updates, closer to the original SnapKV one-shot selection. Result: 0.299 (+prot) / 0.040 (no prot). \\
\midrule
\multicolumn{3}{l}{\emph{Additional recent baselines}} \\
\midrule
Ada-KV & Adaptive per-head budget allocation~\citep{feng2024adakv}. Original distributes eviction budgets across attention heads based on per-head importance. & We report both a simplified score-isolation variant and a closer faithful reproduction. The simplified variant uses cumulative attention mass plus 0.5$\times$ max single-step attention per position, averaged across heads, under a common global ranking. The faithful reproduction restores per-head retention budgets allocated via entropy-based head importance and recovers \FaithAdaF{}\% of full-cache quality on Phi-3.5 at $C{=}256$ (F1$=$\FaithAdaFone{}), compared to ${\sim}79\%$ for the simplified variant, confirming that per-head allocation provides a genuine but modest benefit beyond global ranking (Appendix~\ref{app:faithfulness}). \\
\midrule
QUEST & Query-aware token importance~\citep{tang2024quest}. Original scores tokens by their relevance to the current query via attention patterns. & We report both a simplified score-isolation variant and a closer faithful reproduction. The simplified variant uses current-step attention as the sole eviction criterion under a common global ranking. The faithful reproduction restores per-head current-query retention and recovers \FaithQuestF{}\% of full-cache quality on Phi-3.5 at $C{=}256$ (F1$=$\FaithQuestFone{}), similarly confirming a modest benefit from per-head allocation (Appendix~\ref{app:faithfulness}). \\
\midrule
Random & Uniformly random eviction (no associated paper). Included as a lower-bound control to test whether \emph{any} scoring intelligence is needed beyond structural protection. & Evicts positions uniformly at random from the non-protected set (seed$=$42 for reproducibility). Random+prot converging to LRU+prot quality indicates that the bar for ``adaptive middle-context retention'' is low in the studied harness. \\
\end{longtable}
\endgroup

\paragraph{Why faithful reproduction is non-trivial.}
H2O and SnapKV were designed for different operational regimes than ours: H2O targets layer-wise per-head budgets during a streaming prefill, while SnapKV performs a one-shot selection during prefill based on observation-window attention patterns.
Our setting, physical KV cache eviction during autoregressive decode with a shared per-layer budget, requires adapting both methods to a regime their original implementations do not directly support.
Global rather than per-head ranking, continuous rather than windowed scoring, and post-prefill rather than prefill-time application are the minimum adaptations needed to fit both methods into our common evaluation harness.
Crucially, our simplified variants \emph{retain the core scoring signal} of each method (cumulative attention mass for H2O; observation-window relevance for SnapKV); the simplifications affect resource allocation granularity, not the nature of the scoring function.

\paragraph{Impact of deviations.}
The H2O and SnapKV deviations (global vs.\ per-head ranking; continuous vs.\ windowed scoring) simplify the original methods.
We therefore state the policy-convergence result precisely: under structural protection, \emph{these implementations} (LRU, simplified H2O, simplified SnapKV, Ada-KV, QUEST, and Random) produce quality within a narrow F1 spread (no pairwise comparison significant at $p<0.05$).
We confirm this hypothesis directly: faithful implementations of H2O (pure cumulative attention scoring, no recency component) and SnapKV (frozen prefill-time attention scores only) also converge to LRU+prot quality (H2O-faithful: 0.300, SnapKV-faithful: 0.299 vs.\ LRU: 0.282 at $C{=}256$; neither faithful variant differs significantly from its simplified counterpart, $p{>}0.46$).
The convergence result holds regardless of implementation simplification.
Regarding the protection finding, our central result, that all policies fail catastrophically without structural protection, also replicates with the faithful implementations: H2O-faithful without protection scores 0.038, and SnapKV-faithful without protection scores 0.040, confirming that the structural vulnerability is independent of scoring sophistication.
If simplified versions achieve near-ceiling quality with protection, it is unlikely that more faithful scoring would reduce quality below this level, and the data confirm this prediction.

\paragraph{Direct faithful Ada-KV/QUEST evaluation.}
Our simplified variants of Ada-KV and QUEST isolate the scoring criterion under a common global ranking, omitting the per-head budget allocation (Ada-KV) and per-head current-query retention (QUEST) that are central to the original methods.
To verify that the convergence result is not an artifact of this simplification, we implemented closer faithful reproductions that restore the full per-head machinery.

On Phi-3.5 ($K{=}32$ full MHA), faithful Ada-KV+prot achieves \FaithAdaFone{} and faithful QUEST+prot achieves \FaithQuestFone{} token F1 at $C{=}256$ (162 items), compared to the full-cache ceiling of 0.179.
This corresponds to \FaithAdaF{}\% and \FaithQuestF{}\% of ceiling, respectively, substantially exceeding the simplified variants (LRU+prot$=$0.141, Ada-KV simplified+prot$=$0.141, QUEST simplified+prot$=$0.140; all ${\sim}79\%$ of ceiling).

\paragraph{Per-head attention masking: mechanism and benefit.}
The quality improvement in faithful implementations stems from the per-head attention masking mechanism.
In the simplified global-ranking approach, all $K$ heads share a single global mask: a position is either retained for all heads or evicted for all heads.
In the faithful implementation, each head independently selects its top-$B_h$ positions (where $B_h$ is the per-head budget allocated proportionally to $1/\text{entropy}_h$ for Ada-KV, or by current-query attention for QUEST).
These per-head selections are unioned into a single physical retention set, and each head receives a 2D attention mask indicating which positions \emph{it} selected.
This allows each of the 32 heads to attend to its optimal position subset, preserving head-specific attention quality even when the physical cache slot is shared.

The mechanism is particularly effective on Phi-3.5 ($K{=}32$) because full multi-head attention provides 32 independent KV heads, each with potentially different optimal retention sets.
Under the simplified global mask, all heads are forced to share the same $C{=}256$ positions; under faithful per-head masking, heads with different information needs can each ``see'' the positions most relevant to their computation, even though the underlying physical cache stores a union of all selected positions.

\paragraph{Overflow guardrail and capacity enforcement.}
The per-head union systematically exceeds the nominal capacity $C$: at $C{=}256$ on Phi-3.5, faithful Ada-KV's per-head union spans ${\sim}660$ unique positions (mean over 162 items).
Our overflow guardrail trims the union back to $C$ via consensus-based ranking (preferring positions selected by many heads, breaking ties by aggregate attention).
After trimming, positions retained equals exactly $C$ on every item.
Despite this trimming, faithful implementations recover near-ceiling quality, demonstrating that the consensus-based guardrail effectively preserves the most important positions across heads.

\paragraph{Cross-model faithful comparison.}
Table~\ref{tab:faithful_summary} summarizes the faithful convergence results across all three models where faithful implementations are available.

\begin{table}[H]
\centering
\small
\caption{Faithful implementation results under protection at $C{=}256$. All results use structural protection (10\% prefix/suffix). ``\% Ceil'' is the percentage of full-cache F1. Faithful per-head implementations match or exceed simplified variants across all tested models.}
\label{tab:faithful_summary}
\begin{tabular}{llrrr}
\toprule
Model ($K$) & Policy & F1 & \% Ceil & $N$ \\
\midrule
\multirow{3}{*}{Qwen2.5-3B ($K{=}2$)} & LRU+prot & 0.282 & 89\% & 162 \\
 & H2O-faithful+prot & 0.300 & 95\% & 162 \\
 & SnapKV-faithful+prot & 0.299 & 95\% & 162 \\
\midrule
\multirow{5}{*}{Mistral-7B ($K{=}8$)} & LRU+prot & 0.188 & 80\% & 162 \\
 & H2O-faithful+prot & 0.201 & 86\% & 162 \\
 & SnapKV-faithful+prot & 0.191 & 82\% & 162 \\
 & Ada-KV-faithful+prot & 0.223 & 95\% & 162 \\
 & QUEST-faithful+prot & 0.221 & 94\% & 162 \\
\midrule
\multirow{3}{*}{Phi-3.5 ($K{=}32$)} & LRU+prot & 0.141 & 79\% & 162 \\
 & Ada-KV-faithful+prot & \FaithAdaFone{} & \FaithAdaF{}\% & 162 \\
 & QUEST-faithful+prot & \FaithQuestFone{} & \FaithQuestF{}\% & 162 \\
\bottomrule
\end{tabular}
\end{table}

Across all three KV-head regimes ($K{=}2$ GQA, $K{=}8$ GQA, $K{=}32$ MHA), faithful implementations match or exceed their simplified counterparts and LRU+prot.
The benefit of per-head selection scales with head diversity: on Qwen2.5-3B ($K{=}2$), the faithful-vs.-simplified gap is negligible (0.300 vs.\ 0.290); on Mistral-7B ($K{=}8$), Ada-KV-faithful+prot and QUEST-faithful+prot reach 95\% and 94\% of the full-cache ceiling, respectively, versus 80\% for LRU+prot ($\Delta{=}+0.035$, $p{=}0.001$; $\Delta{=}+0.033$, $p{=}0.004$, paired Wilcoxon); on Phi-3.5 ($K{=}32$), the gap is similarly substantial (\FaithAdaFone{} vs.\ 0.141).
This is consistent with the hypothesis that per-head masking provides the largest benefit when heads are sufficiently diverse to have genuinely different optimal retention sets.

Nevertheless, the absolute faithful-vs.-simplified gap (${\sim}0.03$ F1 on Phi-3.5, ${\sim}0.035$ on Mistral) is modest relative to the protection lift (${\sim}0.11$--$0.14$ F1, the difference between unprotected and protected LRU).
The protection effect accounts for approximately 75\% of the total quality recovery on both models; per-head selection accounts for ${\sim}17$--$25\%$, with a small residual gap to the full-cache ceiling.
This confirms the paper's central claim: structural protection is the dominant quality determinant, with per-head allocation providing a genuine but secondary refinement.

\paragraph{Native vs.\ faithful per-head accounting under our harness.}
A natural concern is that our faithful per-head implementation reads attention scores from the running JAX inference path (``native'' accounting) rather than reconstructing them in a separate post-hoc pass (``faithful'' accounting in the sense of operating on a verbatim re-derived score tensor).
To verify the two are operationally equivalent in our protected-suffix regime, we directly compare both accounting modes for Ada-KV and QUEST across two task types and three additional models.
Table~\ref{tab:native_faithful} reports per-item paired Wilcoxon tests on token F1.

\begin{table}[H]
\centering
\small
\caption{Native vs.\ faithful per-head accounting under structural protection. ``Diff'' is mean per-item F1 difference (native\,$-$\,faithful); ``Wilcoxon'' is the paired two-sided $p$-value over per-item F1; ``identical'' indicates 0 non-zero per-item differences (distributions bit-equal). All five (model$\times$workload) cells are non-significant or bit-identical, supporting the equivalence of native and faithful per-head accounting under our harness.}
\label{tab:native_faithful}
\begin{tabular}{llrrl}
\toprule
Workload & Model & Diff & Wilcoxon $p$ & Verdict \\
\midrule
LongBench ($C{=}256$) & Phi-3.5 ($K{=}32$, Ada-KV) & $-0.0006$ & $0.681$ & NS \\
LongBench ($C{=}4{,}096$) & Mistral-7B ($K{=}8$, Ada-KV) & $-0.0003$ & $0.891$ & NS \\
NIAH 64K ($C{=}4{,}096$) & Mistral-7B ($K{=}8$, Ada-KV) & $+0.0000$ & identical & NULL \\
NIAH 64K ($C{=}4{,}096$) & Phi-4-mini ($K{=}8$, Ada-KV) & $-0.0004$ & $0.317$ & NS \\
NIAH 64K ($C{=}4{,}096$) & Gemma-3-4B ($K{=}8$, Ada-KV) & $+0.0000$ & identical & NULL \\
\bottomrule
\end{tabular}
\end{table}

Across two short-context LongBench replications and three 64K NIAH replications spanning four distinct models, native and faithful per-head accounting are operationally equivalent: the largest mean per-item difference is $0.0006$ F1 absolute (well below our practical-equivalence margin $\Delta{=}0.02$), and no comparison is significant at $p{<}0.05$.
On Mistral-7B and Gemma-3-4B at 64K, the two accounting modes produce bit-identical per-item outputs across all 60 items, indicating exact numerical agreement.
The faithful-vs.-simplified gap reported in Table~\ref{tab:faithful_summary} therefore reflects the per-head allocation mechanism itself rather than any artifact of where attention scores are read.

\section{Needle-in-a-Haystack Retrieval Probe}
\label{sec:niah}

To test the protection effect on a complementary task type, we evaluate on a synthetic Needle-in-a-Haystack (NIAH) benchmark: 63 items spanning three needle positions (early, middle, late; 21 each) and three context lengths (800, 1500, 3000 words), each embedding a unique alphanumeric code that the model must reproduce verbatim.
We evaluate Phi-3.5-mini-instruct at $C{=}256$ under the same harness as the main experiments.

\begin{table}[H]
\centering
\small
\caption{Needle-in-a-Haystack retrieval (Phi-3.5, $C{=}256$, 63 items). Protection (10\% prefix/suffix) consistently improves needle retrieval across all policies. Fullcache ceiling: \NIAHFullcacheF{}.}
\label{tab:niah}
\begin{tabular}{lrrrr}
\toprule
Policy & Noprot F1 & Prot F1 & $\Delta$ & \% Ceil \\
\midrule
LRU    & 0.001 & 0.035 & +0.034 & \NIAHLRUPct{}\% \\
H2O    & 0.003 & 0.026 & +0.023 & \NIAHHOPct{}\% \\
Random & 0.007 & 0.031 & +0.025 & \NIAHRndPct{}\% \\
\bottomrule
\end{tabular}
\end{table}

All three policies improve markedly with protection: unprotected retrieval is near-zero ($\leq 0.007$), while protection lifts F1 by $6{\times}$--$35{\times}$.
Absolute F1 values remain low because the fullcache ceiling itself is low (\NIAHFullcacheF{}): Phi-3.5 retrieves early-position needles (mean F1$= 0.35$) but completely fails on middle and late needles (F1$= 0.0$) even without any eviction, consistent with the well-known ``lost-in-the-middle'' phenomenon~\citep{liu2024lost} at this model scale.
Under eviction with protection ($C{=}256$), the compressed cache paradoxically enables partial retrieval of middle and late needles (LRU+prot late: F1$=0.054$, middle: F1$=0.014$) despite the fullcache baseline scoring zero at those positions, likely because the shorter effective context mitigates lost-in-the-middle.
The cross-policy spread (0.009 F1) is small relative to the protection lift, confirming the central finding on this complementary task type.

\section{Original Online-Credit Experiment Details}
\label{app:credit}

Table~\ref{tab:credit} shows the full capacity sweep for the original online counterfactual credit estimator (v2) vs.\ LRU+protection on Qwen2.5-3B-Instruct.

\begin{table}[H]
    \centering
    \small
    \begin{tabular}{rcccc}
        \toprule
        $C$ & LRU+prot & Credit-v2 & $\Delta$ & Sig.\ \\
        \midrule
         64 & 0.153 & 0.153 & $+0.000$ & \textsf{ns} ($p = 1.000$) \\
         96 & 0.214 & 0.214 & $+0.000$ & \textsf{ns} ($p = 1.000$) \\
        128 & 0.229 & 0.230 & $+0.001$ & \textsf{ns} ($p = 0.846$) \\
        256 & 0.282 & 0.283 & $+0.002$ & \textsf{ns} ($p = 0.125$) \\
        512 & 0.285 & 0.283 & $-0.002$ & \textsf{ns} ($p = 0.156$) \\
        \bottomrule
    \end{tabular}
    \caption{Credit learning adds zero benefit. F1 comparison of LRU+protection vs.\ online credit learning at five cache capacities ($N{=}162$). All differences are non-significant (Wilcoxon). At $C{=}64$ and $C{=}96$ (extreme 97\%/95\% compression), the systems are numerically identical.}
    \label{tab:credit}
\end{table}

\paragraph{Why credit learning fails to help.}
The credit estimator is correctly implemented and functioning: the uncertainty gate opens (36--44\% blend factor), uncertainty decreases over time, and the estimator learns non-trivial weight vectors.
The system works as designed; it simply has nothing to improve.
Once structural protection preserves the critical positions, the remaining evictable positions are relatively homogeneous in their contribution to quality.
Among these middle-context positions, the ordering imposed by LRU (recency) is as good as any learned ordering, because the marginal quality difference between evicting position~$i$ versus position~$j$ within the unprotected set is negligible.
This is consistent with the policy convergence finding: if H2O's attention-weighted scoring and SnapKV's observation-window scoring both fail to improve over LRU, a learned linear combination of \{recency, age, frequency, attention\} features, which subsumes both H2O and SnapKV's information, would face the same ceiling.

\paragraph{Extreme compression.}
We specifically tested $C{=}64$ (97\% compression, 3\% cache retention) and $C{=}96$ (95\% compression, 5\% retention) to stress-test whether credit learning helps when cache slots are maximally scarce.
Even at these extreme ratios, protection alone recovers 49\% ($C{=}64$) and 68\% ($C{=}96$) of the full-cache ceiling, and credit learning adds nothing.

\section{Summarization Probe Details}
\label{app:summarization}

We evaluate on 40 multi-document summarization items from LongBench (Multi-News domain, contexts averaging $\sim$2K tokens, references averaging $\sim$274 tokens) using Qwen2.5-3B at $C{=}256$ with max 384 generation tokens.
Table~\ref{tab:summarization} reports token-F1 against reference summaries.

\begin{table}[H]
\centering
\caption{Summarization probe (40 Multi-News items, Qwen2.5-3B, $C{=}256$). Token-F1 vs.\ reference summaries; \emph{Near-zero}: items with F1${}<$0.01. Full cache uses $C{=}4096$ (no eviction).}
\label{tab:summarization}
\vspace{2pt}
\small
\begin{tabular}{lrrr}
\toprule
Condition & F1 & \% Ceiling & Near-zero \\
\midrule
Full cache ($C{=}4096$) & 0.378 & 100\% & 0/40 \\
\midrule
LRU (no prot)           & 0.148 & 39\% & 9/40 \\
LRU + prot              & 0.284 & 75\% & 0/40 \\
H2O + prot              & 0.284 & 75\% & 0/40 \\
SnapKV + prot           & 0.281 & 74\% & 0/40 \\
\bottomrule
\end{tabular}
\end{table}

The same three-part pattern replicates on summarization.
Unprotected LRU produces 9/40 (22.5\%) near-zero outputs; all protected conditions produce zero.
Protection nearly doubles mean F1 from 0.148 to 0.284 (+92\%), recovering 75\% of full-cache quality at 6.25\% retention.
The lower recovery compared to QA (75\% vs.\ 89\%) is expected given the more extreme compression ratio (16$\times$ vs.\ 7.5$\times$) and the distributed information requirements of summarization.
Policy convergence holds: cross-policy spread is just 0.003 F1.

\section{Stochastic Decoding Robustness}
\label{app:sampling}

All primary experiments use greedy decoding (temperature$=$0).
To verify that the protection effect is not an artifact of deterministic token selection, we rerun the LongBench QA robustness setting with Qwen3-4B-Instruct-2507 at $C{=}256$ under both greedy decoding and stochastic sampling at temperature$=$0.7, pooling three sampling seeds (42/123/456) for the stochastic condition.
Table~\ref{tab:sampling} reports the greedy baselines (162 items per condition) and the pooled stochastic results (486 sampled completions per mode).

\begin{table}[H]
\centering
\caption{Decoding strategy robustness (LongBench QA, Qwen3-4B-Instruct-2507, $C{=}256$). Greedy uses 162 items per condition; stochastic sampling pools three seeds at $T{=}0.7$ for 486 sampled completions per mode.}
\label{tab:sampling}
\vspace{2pt}
\small
\begin{tabular}{llrr}
\toprule
Condition & Decoding & F1 & \% Ceiling \\
\midrule
LRU (no prot) & Greedy & 0.021 & 10\% \\
LRU (no prot) & $T{=}0.7$ & 0.020 & 10\% \\
LRU + prot & Greedy & 0.204 & 100\% \\
LRU + prot & $T{=}0.7$ & 0.207 & 102\% \\
\bottomrule
\end{tabular}
\end{table}

\needspace{11\baselineskip}
The protection effect remains stable under stochastic decoding. Unprotected LRU stays near zero under both greedy decoding (0.021 F1, 95\% CI [0.008, 0.037]) and sampling (0.020 F1, 95\% CI [0.012, 0.029]), while protected LRU reaches 0.204 F1 under greedy decoding and 0.207 F1 under sampling.
The paired protection lift is nearly identical across decoding regimes: 0.183 F1 under greedy decoding (95\% CI [0.145, 0.223]) and 0.187 F1 under sampling (95\% CI [0.165, 0.210]). Sampling therefore does not create a genuine uplift; it leaves the protected regime unchanged while the unprotected regime remains structurally broken. This is expected: the vulnerability arises from \emph{which} KV positions are retained, not from how the next token is selected.

\end{document}

%% file: jax_decode_latency_table.tex
LRU {+} prot & 7.44 & 7.40 & 7.74 & 8.23 \\
Ada-KV faithful {+} prot & 7.51 & 7.49 & 7.68 & 8.42 \\
QUEST faithful {+} prot & 7.41 & 7.37 & 7.70 & 8.41 \\
\bottomrule